\documentclass[lettersize,journal]{IEEEtran}
\usepackage{amsmath,amsfonts}
\usepackage{algorithmic}
\usepackage{algorithm}
\usepackage{array}
\usepackage{textcomp}
\usepackage{stfloats}
\usepackage{url}
\usepackage{verbatim}
\usepackage{graphicx}
\usepackage{caption}
\usepackage{subcaption}
\usepackage{booktabs}
\usepackage{tabularx}
\usepackage[numbers]{natbib}

\usepackage{amssymb}
\usepackage{mathtools}
\usepackage{amsthm}
\usepackage{bm}

\theoremstyle{plain}
\newtheorem{theorem}{Theorem}
\newtheorem{proposition}[theorem]{Proposition}
\newtheorem{lemma}[theorem]{Lemma}

\theoremstyle{definition}

\newtheorem{assumption}[theorem]{Assumption}
\theoremstyle{remark}

\usepackage{hyperref}
\usepackage{cleveref}

\usepackage{xcolor}

\begin{document}

\title{GTNet: A Tree-Based Deep Graph Learning Architecture}


\author{Nan Wu,
	 Chaofan Wang
	\thanks{N. Wu is with the Institute
		for Financial Services Analytics, University of Delaware, DE,
		USA}
	\thanks{C. Wang is an independent researcher}
	\thanks{\copyright~2022 IEEE. Personal use of this material is permitted. Permission from IEEE must be obtained for all other uses, in any current or future media, including reprinting/republishing this material for advertising or promotional purposes, creating new collective works, for resale or redistribution to servers or lists, or reuse of any copyrighted component of this work in other works.}
}




\maketitle

\begin{abstract}
We propose Graph Tree Networks (GTNets), a deep graph learning architecture with a new general message passing scheme that originates from the tree representation of graphs. In the tree representation, messages propagate upward from the leaf nodes to the root node, and each node preserves its initial information prior to receiving information from its child nodes (neighbors). We formulate a general propagation rule following the nature of message passing in the tree to update a node's feature by aggregating its initial feature and its neighbor nodes' updated features. Two graph representation learning models are proposed within this GTNet architecture - Graph Tree Attention Network (GTAN) and  Graph Tree Convolution Network (GTCN), with experimentally demonstrated state-of-the-art performance on several popular benchmark datasets. Unlike the vanilla Graph Attention Network (GAT) and Graph Convolution Network (GCN) which have the "over-smoothing" issue, the proposed GTAN and GTCN models can go deep as demonstrated by comprehensive experiments and rigorous theoretical analysis.
\end{abstract}


\section{Introduction}
\label{intro}
\IEEEPARstart{G}{raph} Neural Networks (GNNs), a class of neural networks for learning on graph structured data, have been successfully applied in many areas to solve real world problems, such as link predictions in social networks \cite{fan2019graph}, pattern recognition in autonomous driving \cite{shi2020point}, product recommendation and personalized search in E-commerce \cite{zhu2019aligraph}, fraud detection in financial services \cite{wang2019semi}, power estimation and tier design in the semiconductor industry \cite{zhang2020grannite, lu2020tp}, traffic forecasting \cite{yu2017spatio}, and natural language processing \cite{yao2019graph, vashishth2020graph, wu2021deep}. Among many different graph representation learning approaches, the class of spatial graph convolution based models, which adopts a message passing scheme to update node features, has gained particular attention due to its simplicity yet good performance. The most representative work among this class is the Graph Convolutional Network (GCN) \cite{DBLP:conf/iclr/KipfW17} which is derived by limiting the ChebNet \cite{defferrard2016convolutional} up to the 1st order polynomial and making approximations on the coefficients and eigenvalue. The success of GCN \cite{ying2018graph, hu2020open} has led to the rapid development in spatial convolution based graph models, such as the Graph Attention Network (GAT) \cite{DBLP:conf/iclr/VelickovicCCRLB18}, GraphSage \cite{hamilton2017inductive}, APPNP \cite{klicpera2018predict}, DAGNN \cite{liu2020towards}, and etc.

Today the vanilla GCN and GAT are the two most popular baseline models. One GCN or GAT layer aggregates only the direct neighbor nodes. Although they work well in many test cases, their performance degrades when stacking multiple propagation layers to achieve larger receptive fields. \cite{xu2018representation} and \cite{li2018deeper} attribute such degradation in model performance to the over-smoothing effect that nodes from different classes become indistinguishable, while \cite{liu2020towards} attributes it to the intertwined propagation and transformation in the models. A small neighborhood may not provide enough information especially when nodes are sparsely labeled \cite{klicpera2018predict, liu2020towards}. Many recent works have been devoted to extending the size of neighborhood utilized in graph learning. APPNP \cite{klicpera2018predict} and DAGNN \cite{liu2020towards} are two most recent deep graph models with state-of-the-art performance. APPNP \cite{klicpera2018predict} is developed based on personalized PageRank to preserve local information which requires fine tuning of a teleport probability. DAGNN \cite{liu2020towards} aggregates neighbors from different hops in parallel (in one layer) to utilize information from a larger receptive field. Although DAGNN has achieved state-of-the-art performance, it may only be applicable to adopt simple aggregation functions such as a mean aggregator. It could be computationally challenging to combine such multi-hop aggregation scheme with attention-based GNN models (such as GAT, and graph transformer \cite{dwivedi2020generalization, hu2020heterogeneous}); because this aggregation scheme requires computation of attention weights for each pair of a node and its k-hop neighbor, which could be a too large number. In the aggregation of k-hop neighborhood, the number of attention weights is the total number of non-zero entries in the k$^{\text{th}}$ power of the graph adjacency matrix $\bm{A}$. Although the adjacency matrix is usually sparse, its k$^{\text{th}}$ power could be non-sparse especially for undirected graphs. For example, in the Cora dataset \cite{wang2019dgl} $\bm{A}^5$ has $\sim$2.2 million non-zero entries while $\bm{A}$ has only $\sim$10k non-zero entries. 

We propose Graph Tree Networks (GTNets), a deep graph learning architecture with a new general message passing scheme that originates from the tree representation of graphs. In the tree representation, each node forms its own tree where the node itself is the root node and all its neighbors up to k-hop are the subnodes. Messages propagate upward from the leaf nodes to the root node in the tree, and each node preserves its initial information prior to receiving information from its child nodes (neighbors). We formulate a general propagation rule following the nature of message passing in the tree to update a node's feature by aggregating its initial feature and its neighbor nodes' updated features. Two graph representation learning models are proposed within this GTNet architecture - Graph Tree Attention Network (GTAN) and Graph Tree Convolution Network (GTCN), with experimentally demonstrated state-of-the-art performance on several popular benchmark datasets. Unlike the vanilla Graph Attention Network (GAT) and Graph Convolution Network (GCN) which have the "over-smoothing" issue \cite{xu2018representation, li2018deeper, chen2020measuring}, the proposed GTAN and GTCN models can go deep by stacking multiple propagation layers.

This work is structured as follows. \Cref{model} illustrates the architecture of the proposed Graph Tree Networks (GTNets). \Cref{GTAN} describes the Graph Tree Attention Network (GTAN) and its comparison with the Graph Attention Network (GAT). \Cref{GTCN} describes the Graph Tree Convolution Network (GTCN) and theoretical analysis on its deep capability, as well as its comparison with the Graph Convolution Network (GCN). \Cref{exp} presents the experimental results demonstrating the superior performance and deep capability of the proposed GTAN and GTCN models. We conclude our work in \Cref{conclusion} and discuss insights for future work.

\section{Graph Tree Networks (GTNets)}
\label{model}

We first introduce notations used throughout this paper. We follow the general convention to use bold uppercase and lowercase to represent matrices and vectors, respectively. The topology of a graph $\mathcal{G}=(\mathcal{V},\mathcal{E})$ with nodes $\mathcal{V}$ and edges $\mathcal{E}$ can be fully described by its adjacency matrix $\bm{A}$ and degree matrix $\bm{D}$. $|\mathcal{V}| = N$ and $|\mathcal{E}| = M$ are the number of nodes and edges, respectively. $\mathcal{N}_u$ denotes the set of direct neighbors (1-hop neighbors) of node $u$. $\bm{X} \in \mathbb{R}^{N \times D}$ represents the feature map for all nodes, where each row $\bm{x}_u \in \mathbb{R}^{1 \times D}$ represents the feature vector of node $u$ with dimension of $D$. $\bm{Y} \in \mathbb{R}^{N \times C}$ represents the class matrix for all nodes, where each row $\bm{y}_u$ represents the class vector of node $u$ with $C$ classes. 

\subsection{Tree Representation and Network Structure} 
\label{GTNet}

\begin{figure}[ht]
	\begin{center}
		\includegraphics[width=\columnwidth]{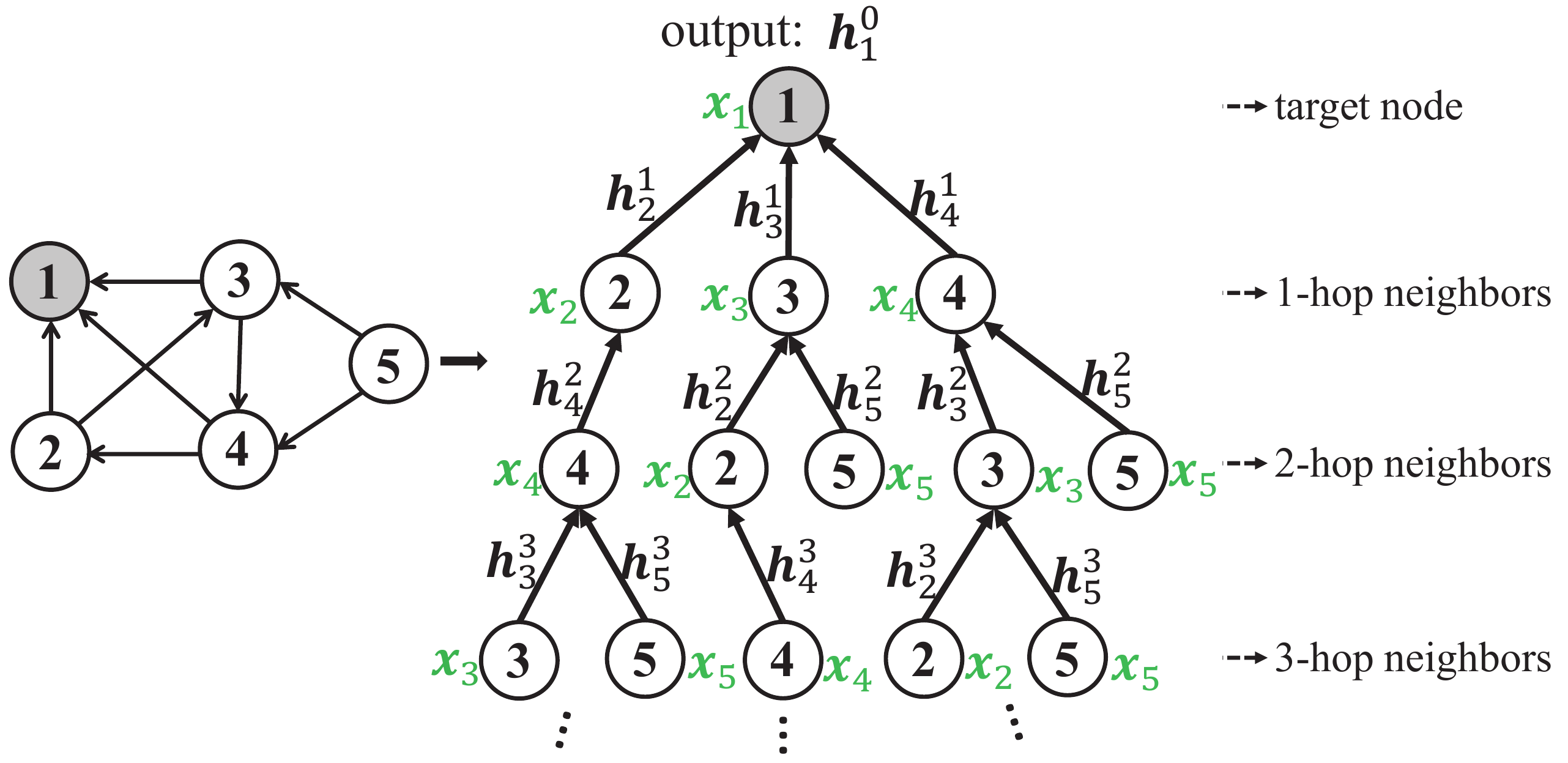}
		\caption{Sample of tree representation for node 1 (showing up to its 3-hop neighborhood). The input feature of node $u$ is denoted as $\bm{x}_u$, and the hidden feature of node $u$ at the $k^{\text{th}}$ hop to the root node is denoted as $\bm{h}_u^k$.}
		\label{fig:tree}
	\end{center}
\end{figure}

Tree is a straightforward representation for the graph topology, where each node and its neighborhood form one tree with the node itself being the root node and its neighbors being the subnodes. Nodes may occur multiple times at the same level in the tree, each of them is from a different k-hop path that may pass different messages to the root node. \Cref{fig:tree} illustrates the tree representation for node 1 in the sample graph with up to its 3-hop neighborhood. We use a directed graph just for illustration, the tree representation also works for undirected graphs.

In \Cref{fig:tree}, $\bm{h}_u^k \in \mathbb{R}^{1 \times F}$ is the hidden feature vector of node $u$ at the k$^\text{th}$ hop to the root node. The final representation of the root node 1 is denoted as $\bm{h}_1^{0}$.

\begin{assumption}
	\label{assumption}
	In a graph tree, messages propagate upward from subnodes to the root node hop by hop in the tree. A node $u$ at any k$^\text{th}$ hop preserves its initial information prior to receiving information from its child nodes (neighbors).
\end{assumption}
With Assumption \ref{assumption}, in the tree representation, a node's feature is updated by aggregating its initial feature and its neighbors' updated features.

Following the nature of message passing in the graph tree as described above, we propose Graph Tree Networks (GTNets) with the following general message passing rule:
\begin{equation}
	\label{eqn:propagation_rule}
	\bm{h}_u^k = f_k\left(\text{aggregate}\left(\phi \left(\bm{x}_u \right), \left\{\bm{h}_v^{k+1}, \forall v \in \mathcal{N}_u \right\}\right)\right)
\end{equation}
where $k=L-1, \ldots, 0$. $L$ is the tree depth (i.e. total number of propagation layers). $\bm{x}_u$ is the input feature vector of node $u$. $\bm{z}_u = \phi(\bm{x}_u)$ is the transformation of input feature of node $u$, i.e. the initial hidden feature for the propagation layer. $\bm{h}_u^k$ is the hidden feature vector of node $u$ at the k$^\text{th}$ hop, and $\bm{h}_u^L = \bm{z}_u$. $f_k(\cdot)$ is a hop-wise (layer-wise) transformation function such as MLP.

\begin{figure}[h]
	\begin{center}
		\includegraphics[width=\columnwidth]{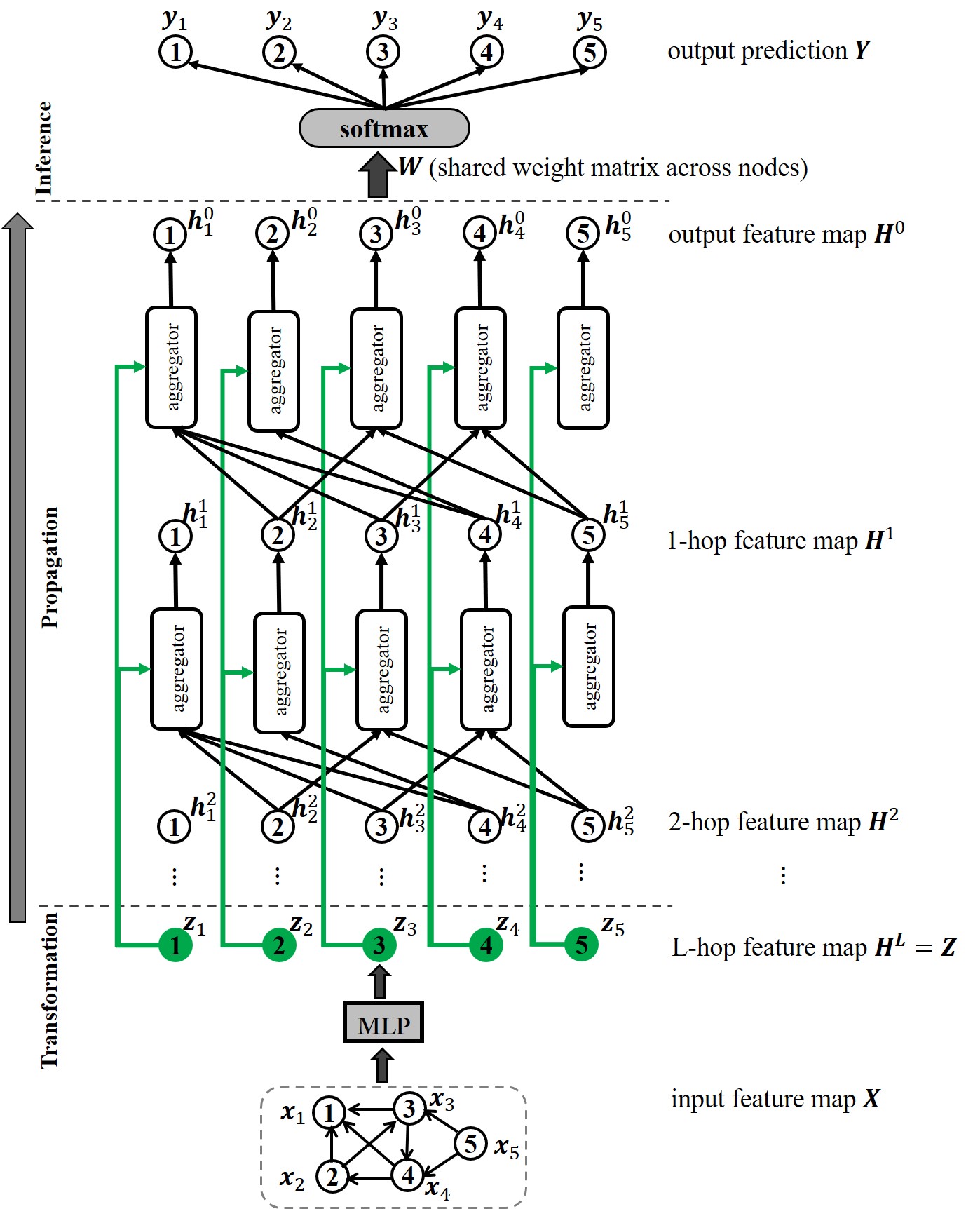}
		\caption{The overall architecture of GTNet on a sample graph.}
		\label{fig:GTNet}
	\end{center}
\end{figure}

\Cref{fig:GTNet} illustrates the overall architecture of GTNet on a sample graph. Green color is used to highlight the initial hidden features of all nodes.

Combination of different aggregation functions and transformation functions in \Cref{eqn:propagation_rule} leads to various graph learning models within the GTNet architecture. Examples of aggregation functions include, but are not limited to, mean aggregator, attention aggregator, weighted sum aggregator, pooling aggregator \cite{hamilton2017inductive}, and LSTM aggregator \cite{hamilton2017inductive}. In this work, we examine an attention aggregator and a modified mean aggregator, and propose two graph tree models in \Cref{GTAN} and \ref{GTCN}: Graph Tree Attention Network (GTAN) and Graph Tree Convolution Network (GTCN).

\subsection{Comparison to Message Passing Scheme in GCN-like models}
\label{compare_rule}
As discussed in \Cref{intro}, a majority of spatial convolution based GNNs is built on the vanilla GCN, such as GAT \cite{DBLP:conf/iclr/VelickovicCCRLB18}, GraphSage \cite{hamilton2017inductive}, and GIN \cite{xu2018powerful}. Their message passing rules can be generalized as\footnote{Note that we use $n$ to denote the number of propagation layers in \Cref{eqn:gcn_rule} to avoid ambiguities. k$^\text{th}$ hop in GTNet corresponds to $(L-k)^\text{th}$ propagation layer in GCN.}:
\begin{equation}
	\label{eqn:gcn_rule}
	\bm{h}_u^n = f_n \left(\text{aggregate}\left( \phi \left( \bm{h}_u^{n-1} \right), \left\{\phi \left(\bm{h}_v^{n-1} \right), \forall v \in \mathcal{N}_u \right\}\right)\right)
\end{equation}
where $\phi\left(\cdot\right)$ is a feature transformation function.

Comparing to \Cref{eqn:propagation_rule}, it is notable that the message passing scheme in the proposed GTNet is essentially different from that in GCN-like models. In GTNet, a node's hidden feature is updated by aggregating its neighbors' output hidden features from the previous propagation layer and its own initial feature. While in GCN-like models, a node's hidden feature is updated by aggregating its neighbors' output hidden features and its own output hidden feature from the previous propagation layer. With Assumption \ref{assumption}, including the focal node's initial feature in the aggregation follows the nature of message passing in the graph tree.

\section{Graph Tree Attention Network}
\label{GTAN}
In this section, we propose a graph tree model - Graph Tree Attention Network (GTAN) which exploits an attention aggregator in GTNet. The message passing rule in GTAN is:
\begin{equation}
	\label{eqn:gtan_node}
	\bm{h}_u^k = \text{ELU}\left( \sum\nolimits_{v\in \mathcal{N}_u} \alpha_{u,v}^k \bm{h}_v^{k+1} + \alpha_{u,u}^k \bm{z}_u \right)
\end{equation}
where $\bm{z}_u = \text{MLP}(\bm{x}_u)$. $\alpha_{u,v}^k$ and $\alpha_{u,u}^k$ are trainable attention weights calculated as
\begin{align}
	\label{eqn:attention_weight}
	\begin{split}
		&\alpha_{u,v}^k = \text{softmax}_v\left(e_{u,v}^k\right) \\
		&e_{u,v}^k =
		\begin{cases}
			\text{LeakyReLU} \left( \left[\bm{z}_u \parallel \bm{h}_v^{k+1}\right] \bm{w}^k \right) & \text{if $u \neq v$}\\
			\text{LeakyReLU} \left( \left[\bm{z}_u \parallel \bm{z}_u\right] \bm{w}^k \right) & \text{if $u = v$}
		\end{cases}
	\end{split}
\end{align}
$\bm{w}^k$ is the trainable layer-wise attention vector.

As compared to the general message passing rule in GTNet in \Cref{eqn:propagation_rule}, our GTAN model uses an MLP as the transformation function $\phi(\cdot)$, an ELU function as $f_k(\cdot)$, and an attention based weighted sum function as the aggregator.

\begin{figure}[h]
	\begin{center}
		\includegraphics[width=\columnwidth]{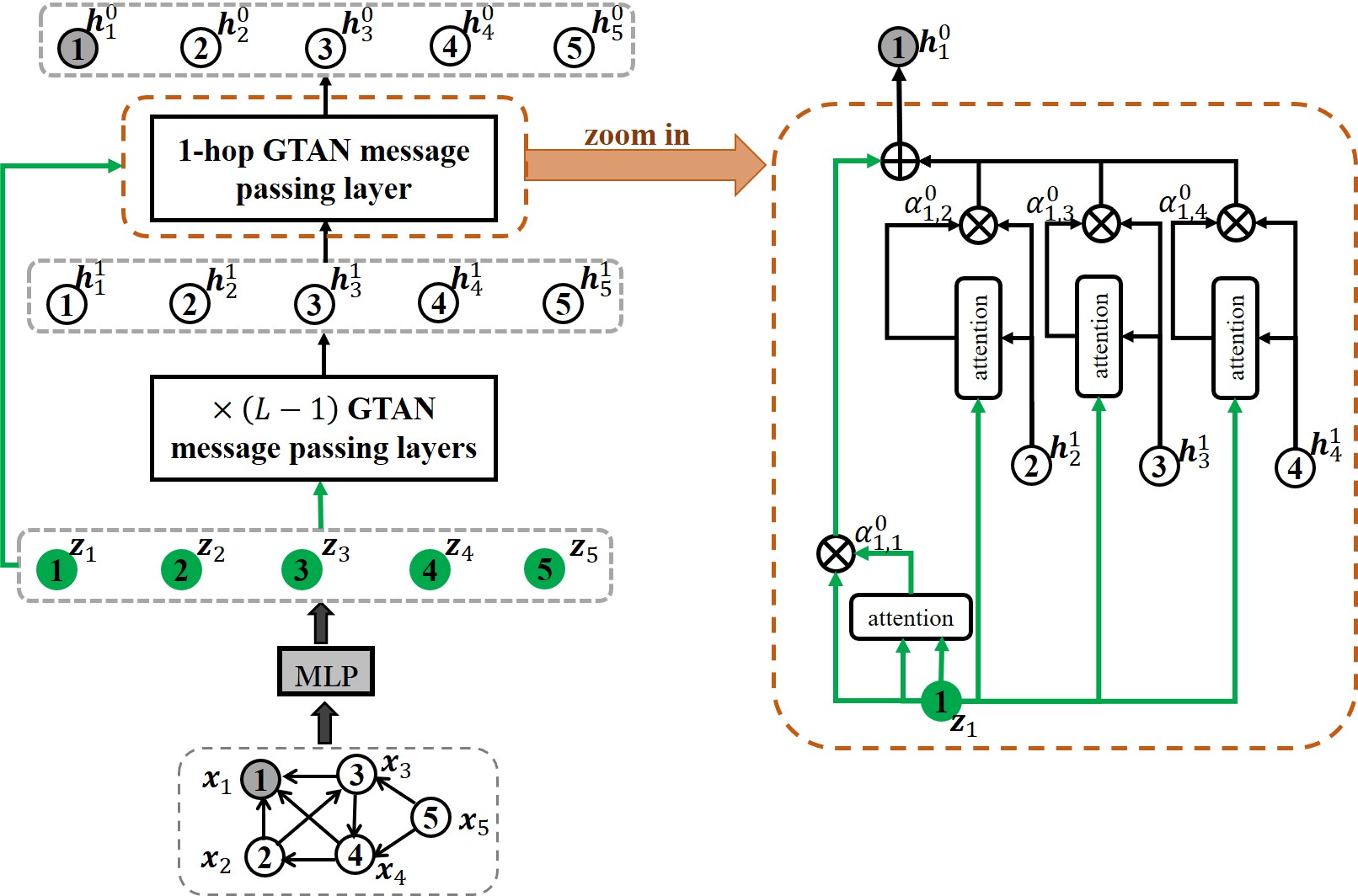}
		\caption{Illustration of GTAN on a sample graph with zoomed in details on the message aggregation at node 1 in one layer.}
		\label{fig:GTAN}
	\end{center}
\end{figure}

\Cref{fig:GTAN} zooms in on the details of the message aggregation at node 1 in one GTAN message passing layer. 

The complete mathematical description of the proposed GTAN model is summarized in below.
\begin{align}
	\label{eqn:GTAN}
	\begin{split}
		&\bm{Z} = \text{MLP}\left(\bm{X}\right) \\
		&\bm{H}^L = \bm{Z} \\
		&\bm{h}_u^k = \text{ELU}\left( \sum\nolimits_{v\in \mathcal{N}_u} \alpha_{u,v}^k \bm{h}_v^{k+1} + \alpha_{u,u}^k \bm{z}_u \right), \\
		&\quad \qquad k = L-1, L-2, \ldots, 0 \\
		&\bm{Y}_{\text{out}} = \text{softmax}\left(\bm{H}^0 \bm{W}_0\right)
	\end{split}
\end{align}
where $\alpha_{u,v}^k$ and $\alpha_{u,u}^k$ are calculated following \Cref{eqn:attention_weight}. $\bm{Z} \in \mathbb{R}^{N \times F}$ is the initial hidden feature map obtained by applying an MLP to the input feature map $\bm{X}$. $\bm{H}^k \in \mathbb{R}^{N \times F}$ is the hidden feature map at the $\text{k}^\text{th}$ hop. $\bm{W}_0 \in \mathbb{R}^{F \times C}$ is the trainable weight matrix.

\subsection{Model Complexity}
The time complexity of our GTAN models is $O(L \lvert \mathcal{E} \rvert F)$ as seen from \Cref{eqn:GTAN}. Here $\lvert \mathcal{E} \rvert$ is the number of edges, $F$ is the dimension of hidden features, and $L$ is the model depth. 

\subsection{Comparison to Graph Attention Network (GAT)}
\label{compare:GAT}
The message passing rule in GAT is
\begin{equation}
	\label{eqn:gat}
	\bm{h}_u^n = \text{ELU} \left( \sum\nolimits_{v\in \mathcal{N}_u} \alpha_{u,v}^n \bm{h}_v^{n-1} \bm{W}^n + \\
	\alpha_{u,u}^n \bm{h}_u^{n-1} \bm{W}^n \right) 
\end{equation}
where
\begin{equation}
	\alpha_{u,v}^n = \text{softmax}_v\left(
	\text{LeakyReLU}\left( \left[\bm{h}_u^{n-1} \bm{W}^n \parallel \bm{h}_v^{n-1} \bm{W}^n \right] \bm{a}^n \right)
	\right)
\end{equation}

In each GAT layer, messages are first transformed and then propagate. The attention weights are calculated with transformed feature vectors. The focal node $u$'s feature is updated by aggregating the transformed hidden features of node $u$ and its neighbors from the previous layer, and followed by an ELU transformation.

In addition to the different feature used in aggregation as discussed in \Cref{compare_rule}, another major difference between our GTAN model and the GAT model is that a parameterized feature transformation is applied in each GAT layer before messages propagate. Like GCN, GAT also suffers from the over-smoothing issue \cite{wang2019improving}. \cite{liu2020towards} argues that intertwined transformation and propagation is the main factor that causes performance degradation in deep GCN. Nevertheless, to our best knowledge, there has no discussion about the causal factor of the over-smoothing in GAT. We experimentally demonstrate that removing the linear feature transformation in GAT does alleviate the over-smoothing problem, but not resolve the over-smoothing problem. We also experimentally demonstrate that introducing the feature transformation in our GTAN model does not jeopardize the model's deep capability.

\section{Graph Tree Convolution Network}
\label{GTCN}
In this section, we propose another graph tree model - Graph Tree Convolution Network (GTCN) which exploits a modified mean aggregator\footnote{The aggregator of GTCN and GCN is actually a weighted sum aggregator. When aggregating neighbors' of node $u$, the weight for $h_v$ is $\frac{1}{\sqrt{d_u +1} \sqrt{d_v + 1}}$} in GTNet. The message passing rule in GTCN is:
\begin{equation}
	\label{eqn:gtcn_node}
	\bm{h}_u^k = \sum\nolimits_{v\in \mathcal{N}_u}{\hat{\bm{A}}_{uv} \bm{h}_v^{k+1}} + \hat{\bm{A}}_{uu} \bm{z}_u
\end{equation}
where $\hat{\bm{A}} = \tilde{\bm{D}}^{-\frac{1}{2}} \tilde{\bm{A}} \tilde{\bm{D}}^{-\frac{1}{2}}$ is the symmetric normalized adjacency matrix\footnote{We find that models using $\hat{\bm{A}} = \tilde{\bm{D}}^{-1}\tilde{\bm{A}}$ show similar performance.}. $\tilde{\bm{A}} = \left(\bm{A} + \bm{I}\right)$ is the adjacency matrix with added self-loops. $\tilde{\bm{D}}$ is the corresponding degree matrix that $\tilde{\bm{D}}_{uu} = \sum\nolimits_v \tilde{\bm{A}}_{uv}$. $\bm{z}_u = \text{MLP}(\bm{x}_u)$.

As compared to the general message passing rule in GTNet shown in \Cref{eqn:propagation_rule}, our GTCN model uses an MLP as the transformation function $\phi(\cdot)$, an identity function as $f_k(\cdot)$, and a modified mean aggregator.

The mathematical description of the proposed GTCN model is then formulated as:
\begin{align}
	\label{eqn:GTCN}
	\begin{split}
		&\bm{Z} =   \text{MLP}\left( \bm{X} \right) \\
		&\bm{H}^L = \bm{Z} \\
		&\bm{H}^k = \bm{A}_1 \bm{H}^{k+1} + \bm{A}_2 \bm{Z}, \; \; k = L-1, L-2, \ldots, 0 \\
		&\bm{Y}_{\text{out}} = \text{softmax}\left(\bm{H}^0 \bm{W}_0\right)
	\end{split}
\end{align}
where $\bm{A}_1$ and $\bm{A}_2$ are the non-diagonal and diagonal components of the normalized adjacency matrix $\hat{\bm{A}}$, respectively.

\begin{theorem}
\label{thm:deep}
	Given an undirected graph $\mathcal{G}=(\mathcal{V,E})$, the node representation learned by a GTCN model with infinite propagation layers converges to the following limit:
	\begin{equation}
		\bm{H}^0_{L \to \infty} = \left(\bm{I} - \bm{A}_1\right)^{-1} \bm{A}_2 \bm{Z}
	\end{equation}
\end{theorem}

To prove Theorem \ref{thm:deep}, we first introduce the following two lemmas.

\begin{lemma}
	\label{lm:eigen1}
	Given an undirected graph $\mathcal{G}=(\mathcal{V,E})$, $\bm{A}$ is the adjacency matrix of $\mathcal{G}$, $\hat{\bm{A}} = \tilde{\bm{D}}^{-\frac{1}{2}} \tilde{\bm{A}} \tilde{\bm{D}}^{-\frac{1}{2}}$ is the symmetric normalized adjacency matrix. $\tilde{\bm{A}} = \left(\bm{A} + \bm{I}\right)$ is the adjacency matrix with added self-loops. $\tilde{\bm{D}}$ is the corresponding degree matrix that $\tilde{\bm{D}}_{uu} = \sum\nolimits_v \tilde{\bm{A}}_{uv}$. $\bm{A}_1$ is the non-diagonal component of $\hat{\bm{A}}$, i.e. 
	$\bm{A}_{1,uv} = \begin{cases}
		\hat{\bm{A}}_{uv} & \text{if $u \neq v$}\\
		0 & \text{if $u = v$}
	\end{cases}$. Let $a_1, a_2, \ldots, a_N$ be the eigenvalues of $\bm{A}_1$, the eigenvalues are bounded that $-1<a_i<1$, where $i=1,\ldots,N$.
\end{lemma}

\begin{lemma}
	\label{lm:eigen2}
	Let $\lambda_1, \lambda_2, \ldots, \lambda_N$ be the eigenvalues of the matrix $\bm{I}-\bm{A}_1$, they are bounded that $0<\lambda_i <2$, where $i=1, \ldots, N$.
\end{lemma}

See Appendix \ref{proof} for proofs of Lemma \ref{lm:eigen1} and \ref{lm:eigen2}, and Theorem \ref{thm:deep}.

\Cref{thm:deep} tells that the proposed GTCN model can go deep without the over-smoothing issue.

\subsection{Model Complexity}
The time complexity of our GTCN models is $O(L \lvert \mathcal{E} \rvert F)$ as seen from \Cref{eqn:GTCN}. Here $\lvert \mathcal{E} \rvert$ is the number of edges, $F$ is the dimension of hidden features, and $L$ is the model depth. 

\subsection{Comparison to Graph Convolution Network (GCN)}
\label{compare:GCN}
The message passing rule in GCN is:
\begin{equation}
	\label{eqn:gcn}
	\bm{h}_u^n = \text{ReLU}\left(\sum\nolimits_{v\in \mathcal{N}_u}{\hat{\bm{A}}_{uv} \bm{h}_v^{n-1}} \bm{W}^n + \hat{\bm{A}}_{uu} \bm{h}_u^{n-1}\bm{W}^n\right)
\end{equation}

In each GCN layer, messages are first transformed and then propagate. The focal node $u$'s feature is updated by aggregating the transformed hidden features of itself and its neighbors from the previous layer with a modified mean aggregator, and followed by an ReLU transformation.

\begin{proposition}
	\label{prop:deep}
	Given a fully connected graph $\mathcal{G}=(\mathcal{V}, \mathcal{E})$, representations learned by GCN are identical for all nodes regardless of the model depth. A GTCN model with more than one layer learns distinguishable representations for different nodes.
\end{proposition}

See Appendix \ref{proof} for proof of Proposition \ref{prop:deep}.

In addition to the different information used in aggregation as discussed in \Cref{compare_rule}, another major difference between our GTCN model and the GCN model is that a parameterized feature transformation is applied in each GCN layer before messages propagate. \cite{liu2020towards} argues that such intertwined transformation and propagation is the main factor that causes performance degradation when stacking multiple GCN layers. However, we experimentally demonstrate that excluding the linear feature transformation in GCN only slows down the over-smoothing problem, while including feature transformation in our GTCN model does not introduce the over-smoothing problem.

\section{Experiments}
\label{exp}

Our experiments are conducted in three parts. The first part in \Cref{exp_GTC_GTAN} is to demonstrate the effectiveness and efficiency of the proposed GTAN and GTCN models. We choose a depth of 10 for both proposed models to aggregate information from up to 10-hop neighborhood, and compare their performance against five baseline models on five popular benchmark datasets. The baseline models include two of the most fundamental GNN models - the vanilla GCN and GAT, two up-to-date best deep GNN models - APPNP and DAGNN, and the Child-Sum Tree-LSTM model \cite{tai2015improved}. The second part in \Cref{exp_depth} is to demonstrate the deep capability of the proposed models by comparing our model performance with the vanilla GCN and GAT at different model depths. The third part in \Cref{ablation} is to discuss the effect of linear feature transformation on models' deep capability.

\textbf{Datasets}. We work on semi-supervised node classification tasks with four popular benchmark datasets (Cora, Citeseer, PubMed and MS Coauthor-CS) from the DGL library \cite{wang2019dgl} and one OGB dataset ogbn-arxiv \cite{hu2020ogb}. The four datasets from the DGL library are undirected graphs, among which Cora, Citeseer, and PubMed are citation network datasets, and Coauthor-CS is a co-authorship graph dataset. We use fixed data splits for the Cora, Citeseer and PubMed datasets which are provided by the DGL library. For Coauthor-CS, we split the data in the following way. We randomly selected 20 training samples per class and 30 validation samples per class, and all remaining samples are used as the test samples. We repeat the above process three times to get three different splits for the Coauthor-CS dataset to avoid any bias. The ogbn-arxiv dataset is a homogeneous, unweighted, and directed graph representing the citation network of arXiv papers in Computer Science indexed by MAG \cite{wang2020microsoft}. Fixed data split for ogbn-arxiv is provided by OGB \cite{hu2020ogb}.

The statistics of the benchmark datasets are summarized in \Cref{tb:datasets}. Note that the adjacency matrices of the Citeseer and PubMed datasets include several self-loops. We remove these self-loops before proceeding.

\begin{table*}[t]
	\caption{Statistics of datasets.}
	\label{tb:datasets}
	\begin{center}
			\begin{tabular}{llllll}
				\toprule
				\textbf{} &\textbf{Cora} &\textbf{Citeseer} &\textbf{PubMed} &\textbf{Coauthor-CS} &\textbf{ogbn-arxiv}\\
				\midrule
				\textbf{\# Nodes}           &2708  &3327  &19717  &18333 &169343\\
				\textbf{\# Edges}           &10556  &9228  &88651  &81894 &1166243\\
				\textbf{\# Features/Node}   &1433 &3703  &500  &6805  &128\\
				\textbf{\# Classes}         &7 &6  &3  &15  &40\\
				\textbf{Label Rate}      &5.2\%  &3.6\%  &0.3\%  &1.6\%  &53.7\%\\
				\textbf{\# Training Nodes}        &20/class &20/class  &20/class  &20/class  &90941\\
				\textbf{\# Validation Nodes}      &500  &500  &500  &30/class   &29799\\
				\textbf{\# Test Nodes}            &1000  &1000  &1000  &17583  &48603\\
				\bottomrule
			\end{tabular}
	\end{center}
\end{table*}

\begin{table*}[t]
	\caption{Average classification accuracy $\pm$ one standard deviation (in percent) on DGL datasets with the top and bottom 10\% data excluded.}
	\label{tb:result1}
	\begin{center}
		\begin{tabularx}{0.9\textwidth}{p{3.5cm} p{1.5cm} p{1.5cm} p{1.5cm} p{1.5cm} p{1.5cm} p{1.5cm}}
			\toprule
			\textbf{Method}  &\textbf{Cora} &\textbf{Citeseer}
			&\textbf{PubMed} &\textbf{Coauthor-CS (1)}
			&\textbf{Coauthor-CS (2)}
			&\textbf{Coauthor-CS (3)} \\
			\midrule
			GCN (hop = 2)  &81.5 $\pm$ 0.2 &71.6 $\pm$ 0.3 &79.0 $\pm$ 0.3 &90.7 $\pm$ 0.2 &90.2 $\pm$ 0.2 &89.7 $\pm$ 0.2\\
			GAT (hop = 2) &83.0 $\pm$ 0.5 &71.1 $\pm$ 0.9 &77.6 $\pm$ 0.4 &90.4 $\pm$ 0.3 &90.3 $\pm$ 0.5 &89.5 $\pm$ 0.4\\
			Tree-LSTM (hop = 2) &81.9 $\pm$ 0.7 &68.6 $\pm$ 1.0 &76.6 $\pm$ 0.6 &91.0 $\pm$ 0.2 &90.8 $\pm$ 0.2 &90.8 $\pm$ 0.2\\
			APPNP (hop = 10) &83.6 $\pm$ 0.5 &71.6 $\pm$ 0.5 &79.6 $\pm$ 0.2 &91.8 $\pm$ 0.4 &91.8 $\pm$ 0.3 &91.4 $\pm$ 0.3\\
			DAGNN (hop = 10) &84.0 $\pm$ 0.5 &72.6 $\pm$ 0.5 &\textbf{79.6} $\pm$ \textbf{0.4} &90.4 $\pm$ 0.3 &90.3 $\pm$ 0.4 &89.6 $\pm$ 0.6\\
			\midrule
			GTAN (hop = 10, ours) &83.7 $\pm$ 0.6 &72.3 $\pm$ 0.7 &79.6 $\pm$ 0.3 &92.3 $\pm$ 0.2 &\textbf{92.5} $\pm$ \textbf{0.3} &92.0 $\pm$ 0.3\\
			GTCN (hop = 10, ours) &\textbf{84.5} $\pm$ \textbf{0.6} &\textbf{72.9} $\pm$ \textbf{0.5} &79.2 $\pm$ 0.3 &\textbf{92.7} $\pm$ \textbf{0.1} &92.5 $\pm$ 0.1 &\textbf{92.4} $\pm$ \textbf{0.2}\\
			\bottomrule
		\end{tabularx}
	\end{center}
\end{table*}

\subsection{Performance of GTAN and GTCN}
\label{exp_GTC_GTAN}

\textbf{Baseline models}. We compare our GTCN and GTAN models to five GNN models: the vanilla GCN \citep{DBLP:conf/iclr/KipfW17}, GAT \citep{DBLP:conf/iclr/VelickovicCCRLB18}, Child-Sum Tree-LSTM \citep{tai2015improved}, APPNP \citep{klicpera2018predict}, and DAGNN \citep{liu2020towards}.

\textbf{Model hyperparameters}. For fair comparisons, all models are set up with nearly the same settings as in their original papers. All implementations use Adam optimizer with two hyperparameters: learning rate and weight decay.

The detailed experimental setup is described in Appendix \ref{exp_setup}.

\textbf{Results on accurancy}. We take 30 runs of each model on each DGL dataset, and 10 runs of each model on the obgn-arxiv dataset\footnote{This is set by OGB.}. The test results on the classification accuracy are represented by mean $\pm$ std. Test results on the four DGL datasets are summarized in \Cref{tb:result1}. Test results on the obgn-arxiv dataset are summarized in \Cref{tb:result_ogb}. We also provide the Macro-F1 scores on the four DGL datasets as summarized in \Cref{tb:F1} in Appendix \ref{result:F1}.

\begin{table}[ht]
	\caption{Average classification accuracy $\pm$ one standard deviation (in percent) on ogbn-arxiv dataset.}
	\label{tb:result_ogb}
	\begin{center}
			\begin{tabular}{ll}
				\toprule
				\textbf{Method} &\textbf{ogbn-arxiv}\\
				\midrule
				GCN (hop = 3)  &71.7 $\pm$ 0.2 \\
				GAT (hop = 3) &71.8 $\pm$ 0.1 \\
				Tree-LSTM (hop = 3) &71.4 $\pm$ 0.3 \\
				APPNP (hop = 5) &71.5 $\pm$ 0.1 \\
				DAGNN (hop = 16) &72.1 $\pm$ 0.3 \\
				\midrule
				GTAN (hop = 4, ours) &\textbf{73.0} $\pm$ \textbf{0.2}\\
				GTCN (hop = 5, ours) &72.3 $\pm$ 0.2\\
				\bottomrule
			\end{tabular}
	\end{center}
\end{table}

Our GTCN model outperforms the current state-of-the-art baseline models on the Cora, Citeseer, and Coauthor-CS datasets. Our GTAN model achieves state-of-the-art performance on the Pubmed, Coauthor-CS and ogbn-arxiv dataset.

The ogbn-arxiv dataset is much larger with more complex neighborhood structures as compared to the four DGL datasets. This may explain why GTAN outperforms GTCN on ogbn-arxiv. GTAN has trainable pair-wise attention weights in each propagation layer, while GTCN does not have trainable weights in all propagation layers, i.e. GTCN employs fixed weights in the message aggregation. Therefore GTAN could capture more complex neighborhood structure than GTCN, and is more favored in such case.

\textbf{Training time per epoch}. The average training time per epoch on the DGL datasets are summarized in \Cref{tb:time}. Our GTAN and GTCN models have similar runtime as GAT and GCN, respectively.

\begin{table*}[t]
	\caption{Average training time per epoch (ms/epoch) on DGL datasets.}
	\label{tb:time}
	\begin{center}
			\begin{tabular}{lllll}
				\toprule
				\textbf{Method} &\textbf{Cora} &\textbf{Citeseer} &\textbf{PubMed} &\textbf{Coauthor-CS} \\
				\midrule
				GCN (hop = 2) &3.5  &3.3  &3.6   &7.4\\
				GAT (hop = 2) &4.7  &4.6  &5.0  &12.0\\
				GCN (hop = 10) &14.3  &13.9  &19.1   &25.9\\
				GAT (hop = 10) &17.6  &17.8  &21.8  &32.7\\
				APPNP (hop = 10) &3.7  &3.5  &3.8  &12.0\\
				DAGNN (hop = 10) &4.7  &4.6  &5.1  &13.2\\
				\midrule
				GTAN (hop = 10, ours) &17.7  &18.1  &19.4  &34.4\\
				GTCN (hop = 10, ours) &3.7  &3.6  &8.1  &18.6\\
				\bottomrule
			\end{tabular}
	\end{center}
\end{table*}

\subsection{Deep Capability of GTAN and GTCN}
\label{exp_depth}

In this section, we demonstrate the deep capability of the proposed GTAN and GTCN models by testing our model performance against the vanilla GAT and GCN with various model depths (i.e. various number of propagation layers) on the four DGL datasets. We first compare the average classification accuracy of the proposed GTAN and GTCN models against the vanilla GAT and GCN at depths of 2, 5, and 10, respectively. The average classification accuracy is obtained with the same process as described in \Cref{exp_GTC_GTAN}. We then compare the training accuracy and validation accuracy over the number of epochs of GTAN and GTCN with GAT and GCN at different depths to take a deeper look into the over-smoothing phenomenon in GAT and GCN.

\begin{figure}[ht]
	\begin{subfigure}[b]{0.235\textwidth}
		\begin{center}
			\includegraphics[width = 1.0\linewidth]{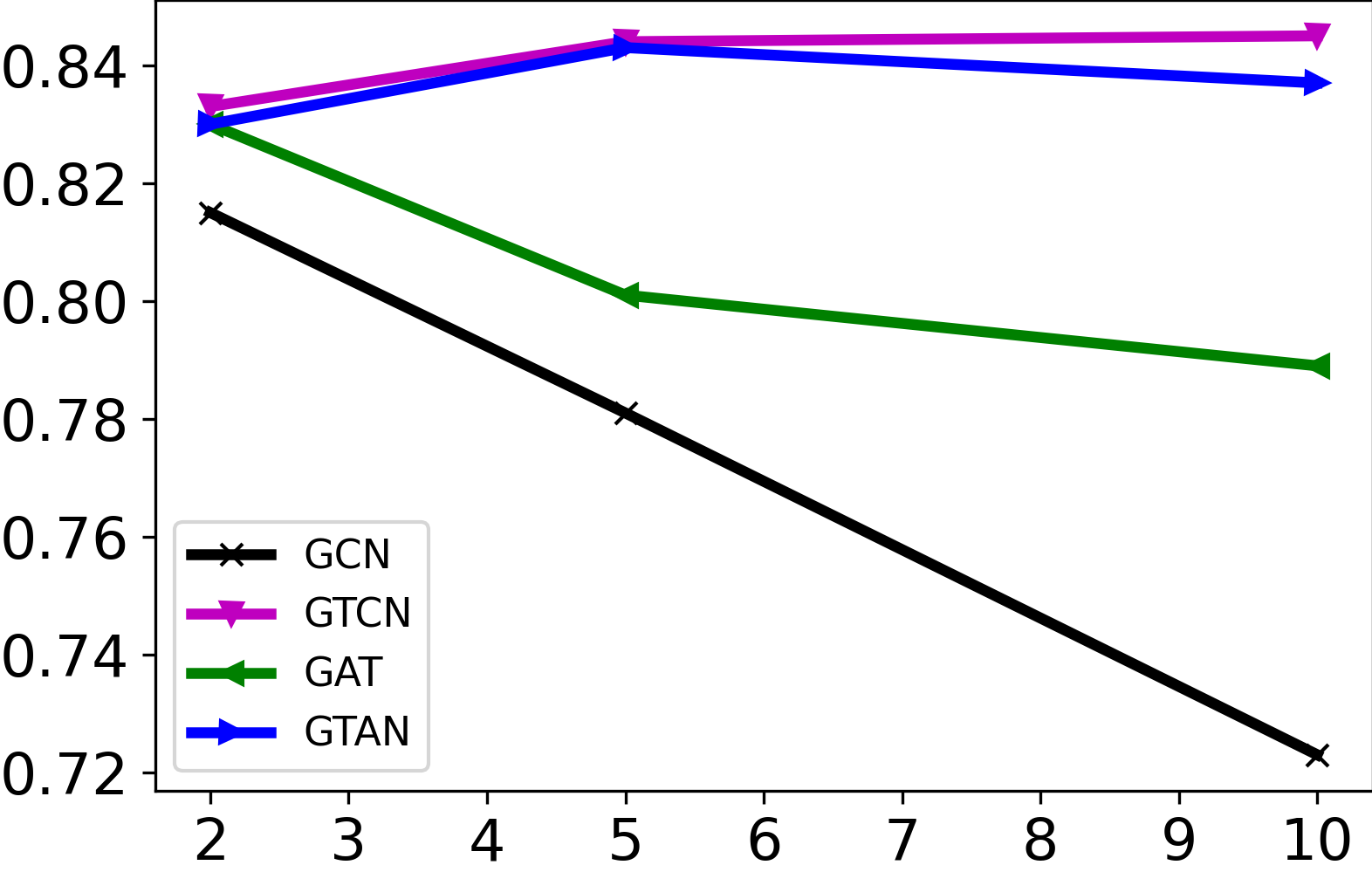}
			\caption{Cora}
			\label{fig:depth_cora}
		\end{center}
	\end{subfigure}
	\begin{subfigure}[b]{0.235\textwidth}
		\begin{center}
			\includegraphics[width = 1.0\linewidth]{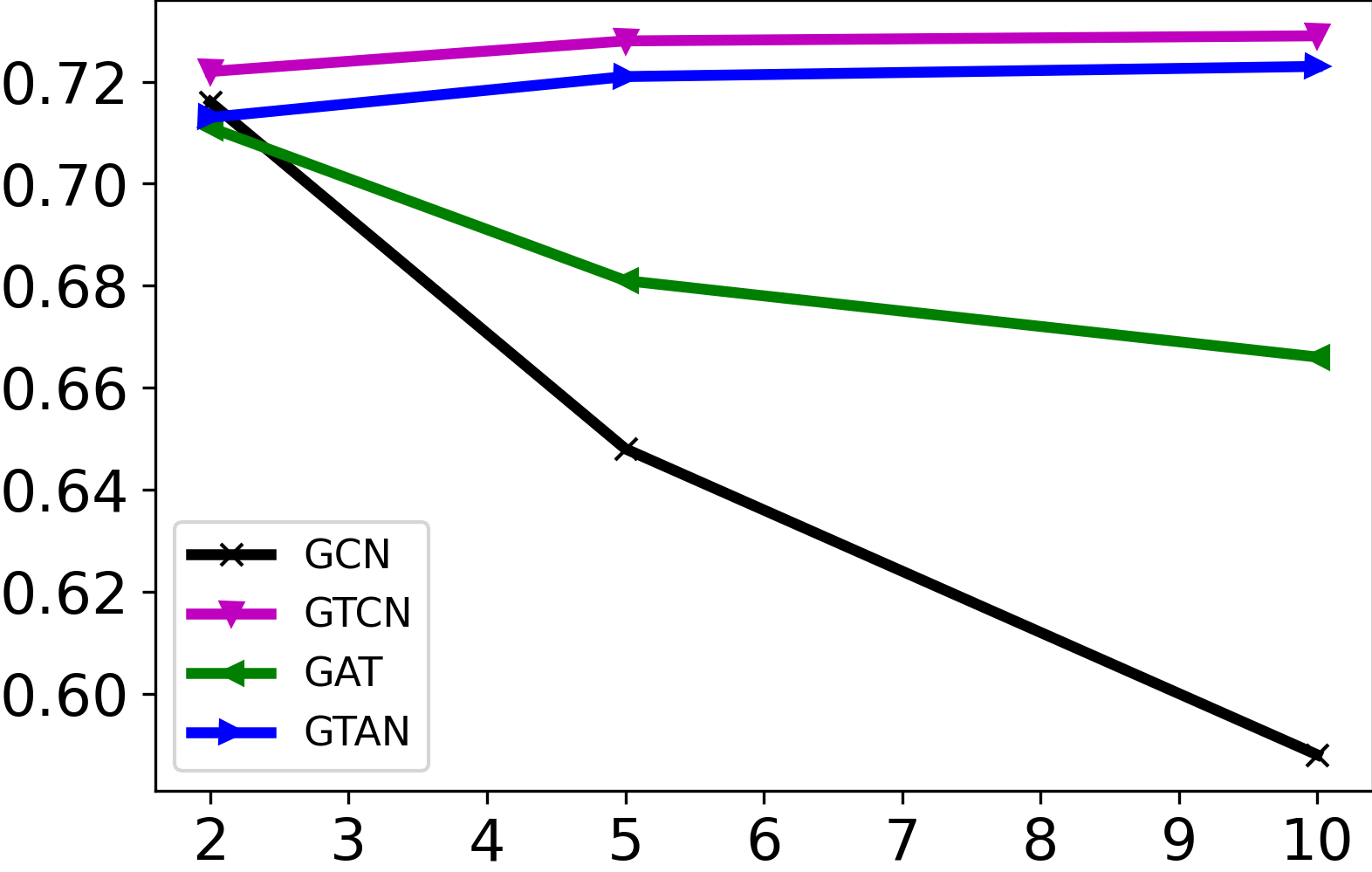}
			\caption{Citeseer}
			\label{fig:depth_citeseer}
		\end{center}
	\end{subfigure}
	\newline
	\begin{subfigure}[b]{0.235\textwidth}
		\begin{center}
			\includegraphics[width = 1.0\linewidth]{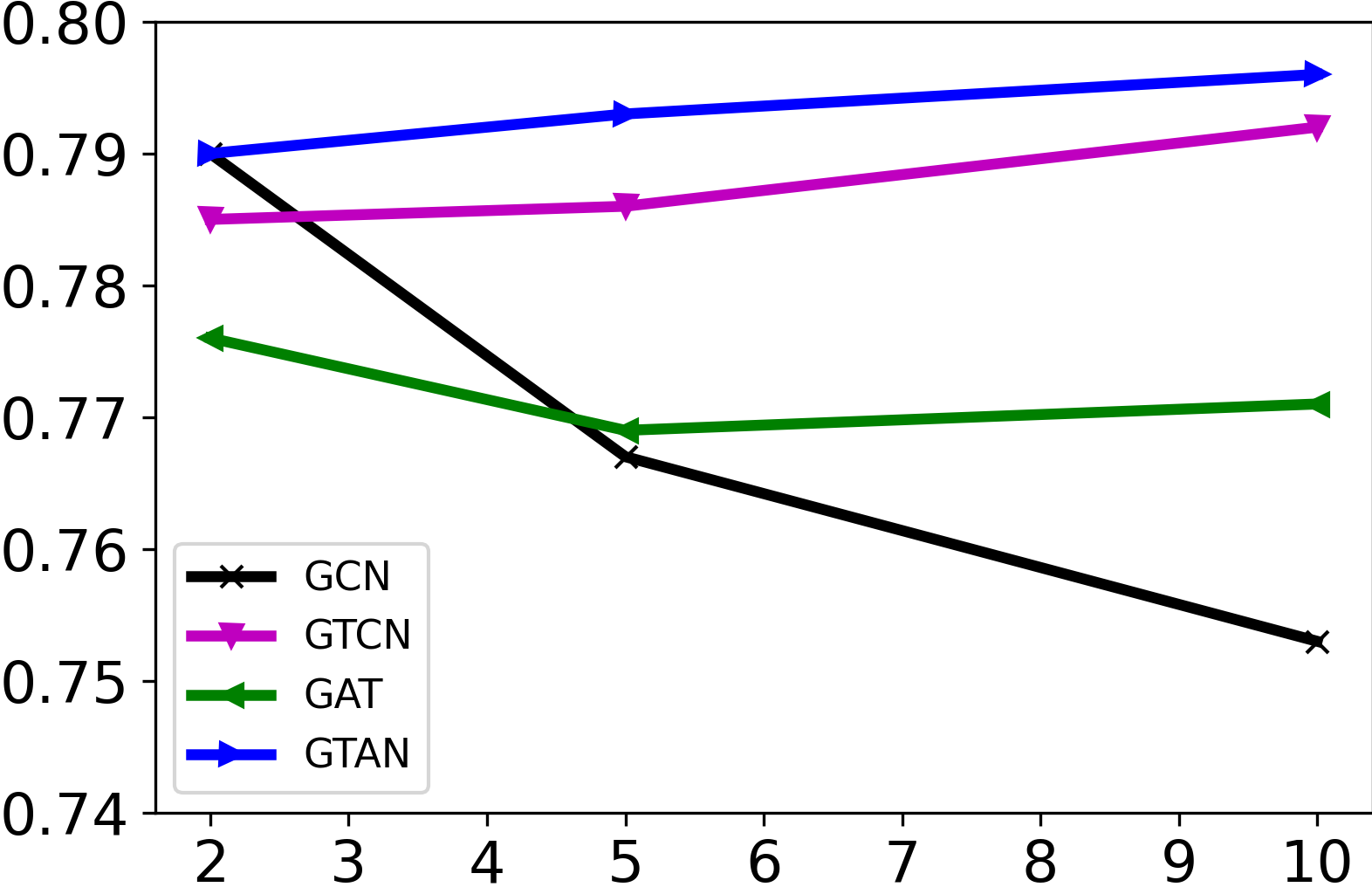}
			\caption{Pubmed}
			\label{fig:depth_pubmed}
		\end{center}
	\end{subfigure}
	\begin{subfigure}[b]{0.235\textwidth}
		\begin{center}
			\includegraphics[width = 1.0\linewidth]{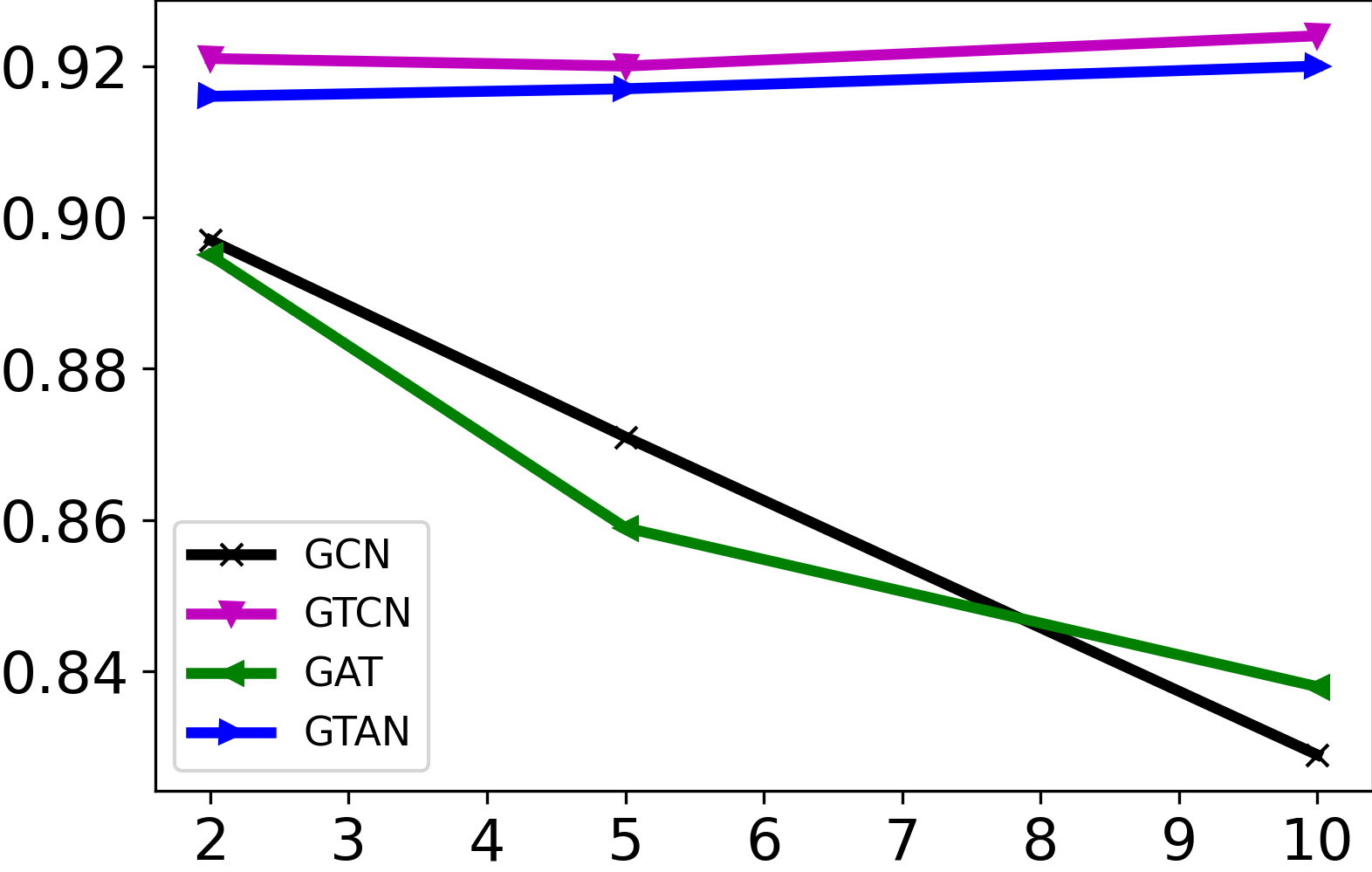}
			\caption{Coauthor-CS}
			\label{fig:depth_coauthor}
		\end{center}
	\end{subfigure}
	\caption{Model performance in terms of average classification accuracy at different depths, tested on four datasets: Cora (a), Citeseer (b), Pubmed (c), and Coauthor-CS (d).}
	\label{fig:depth}
\end{figure}

\textbf{Model hyperparameters}. For all models under test, we use 64 hidden units for every intermediate hidden layer. For the vanilla GCN, the dropout is 0.5, learning rate is 0.01 and weight decay is 5e-4 for all four datasets and all model depths. For the vanilla GAT, we use one attention head for all experiments. At depth 2, the hyperparameters are the same as used in \Cref{exp_GTC_GTAN}. At depth 5, the hyperparameters are mostly the same as used at depth 2 except that we set 0 for the attention dropout for all four datasets. At depth 10, the hyperparameters are fine-tuned with the layer dropout and the attention dropout of (0.2, 0) for all four datasets. The learning rate and weight decay are 0.01 and 5e-3 respectively for all four datasets. For our GTCN model, the settings are the same as used in \Cref{exp_GTC_GTAN}. For our GTAN model, the settings for depth 2 and 5 are mostly the same as described in \Cref{exp_GTC_GTAN}, except that the attention dropout is set to 0.6 for the Cora and Citeseer datasets.

\textbf{Results}. The average classification accuracy of all four models at different depths are displayed in \Cref{fig:depth}. The results show that the performance of our GTAN and GTCN models does not compromise when models going deep, while the performance of the vanilla GAT and GCN degrades significantly when going deep which is known as the "over-smoothing" issue.

\begin{figure}[ht]
	\begin{subfigure}[b]{0.235\textwidth}
		\begin{center}
			\includegraphics[width = 1.0\linewidth]{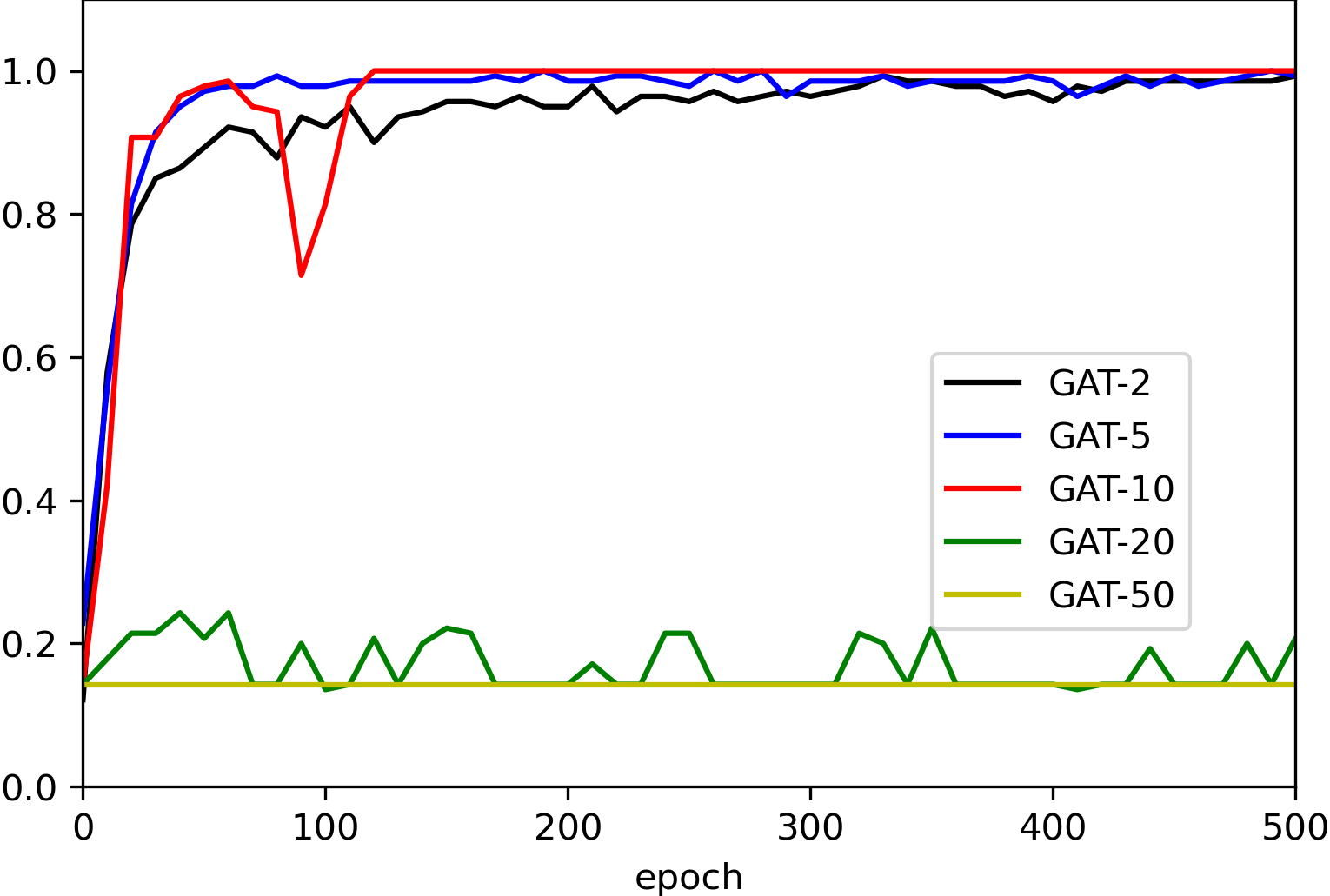}
			\caption{GAT Training Accuracy}
			\label{fig:GAT_train}
		\end{center}
	\end{subfigure}
	\begin{subfigure}[b]{0.235\textwidth}
		\begin{center}
			\includegraphics[width = 1.0\linewidth]{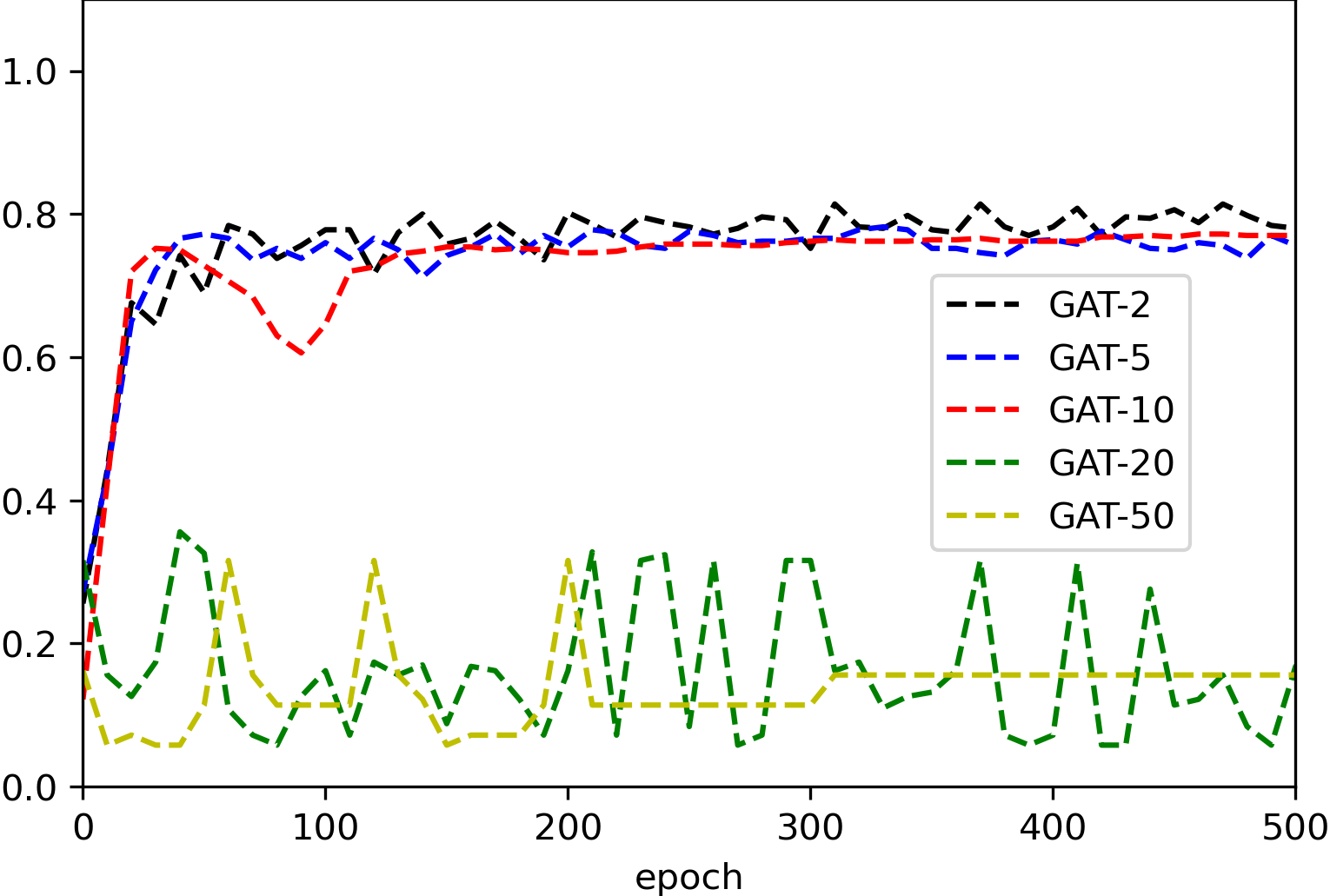}
			\caption{GAT Validation Accuracy}
			\label{fig:GAT_val}
		\end{center}
	\end{subfigure}
	\newline
	\begin{subfigure}[b]{0.235\textwidth}
		\begin{center}
			\includegraphics[width = 1.0\linewidth]{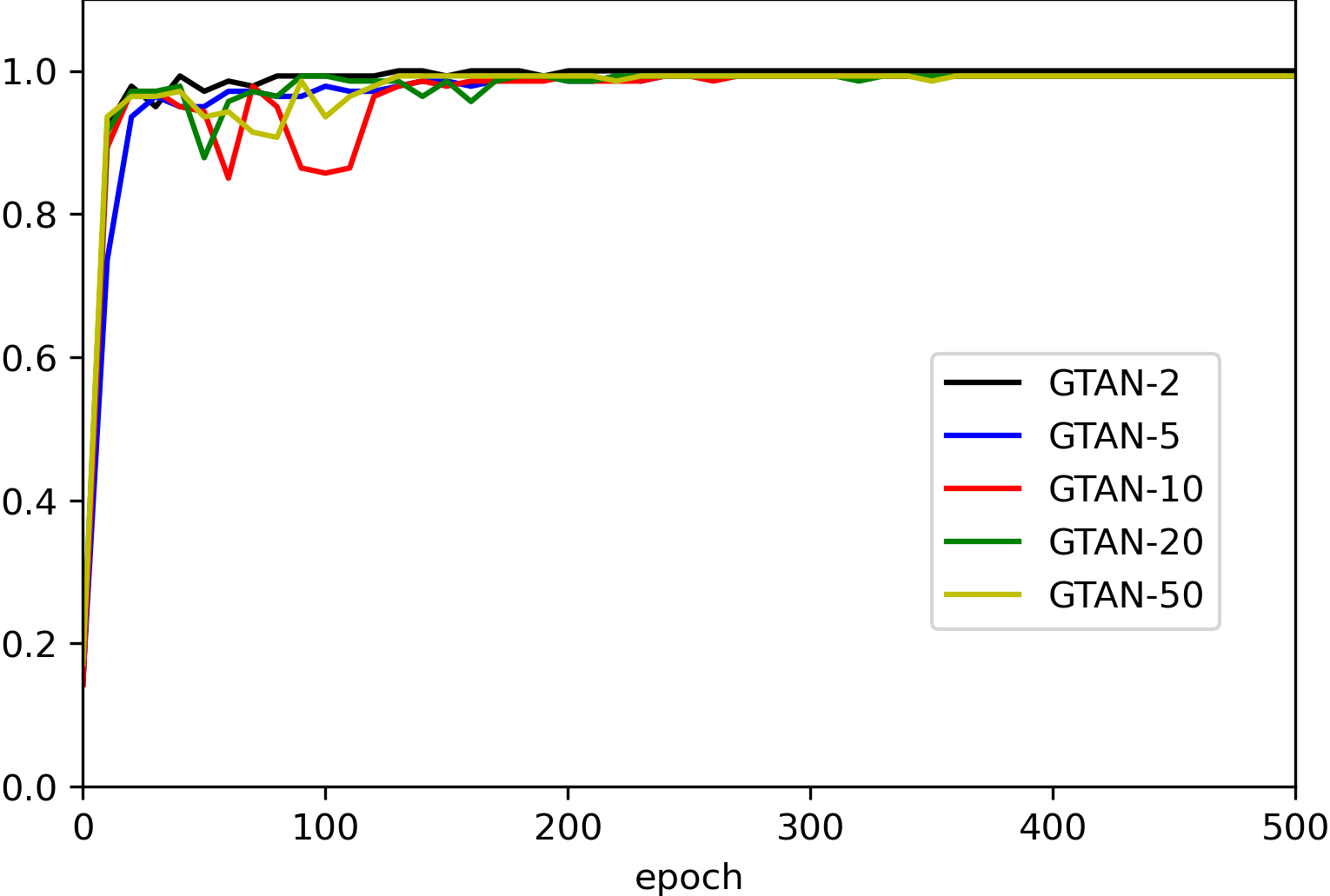}
			\caption{GTAN Training Accuracy}
			\label{fig:GTAN_train}
		\end{center}
	\end{subfigure}
	\begin{subfigure}[b]{0.235\textwidth}
		\begin{center}
			\includegraphics[width = 1.0\linewidth]{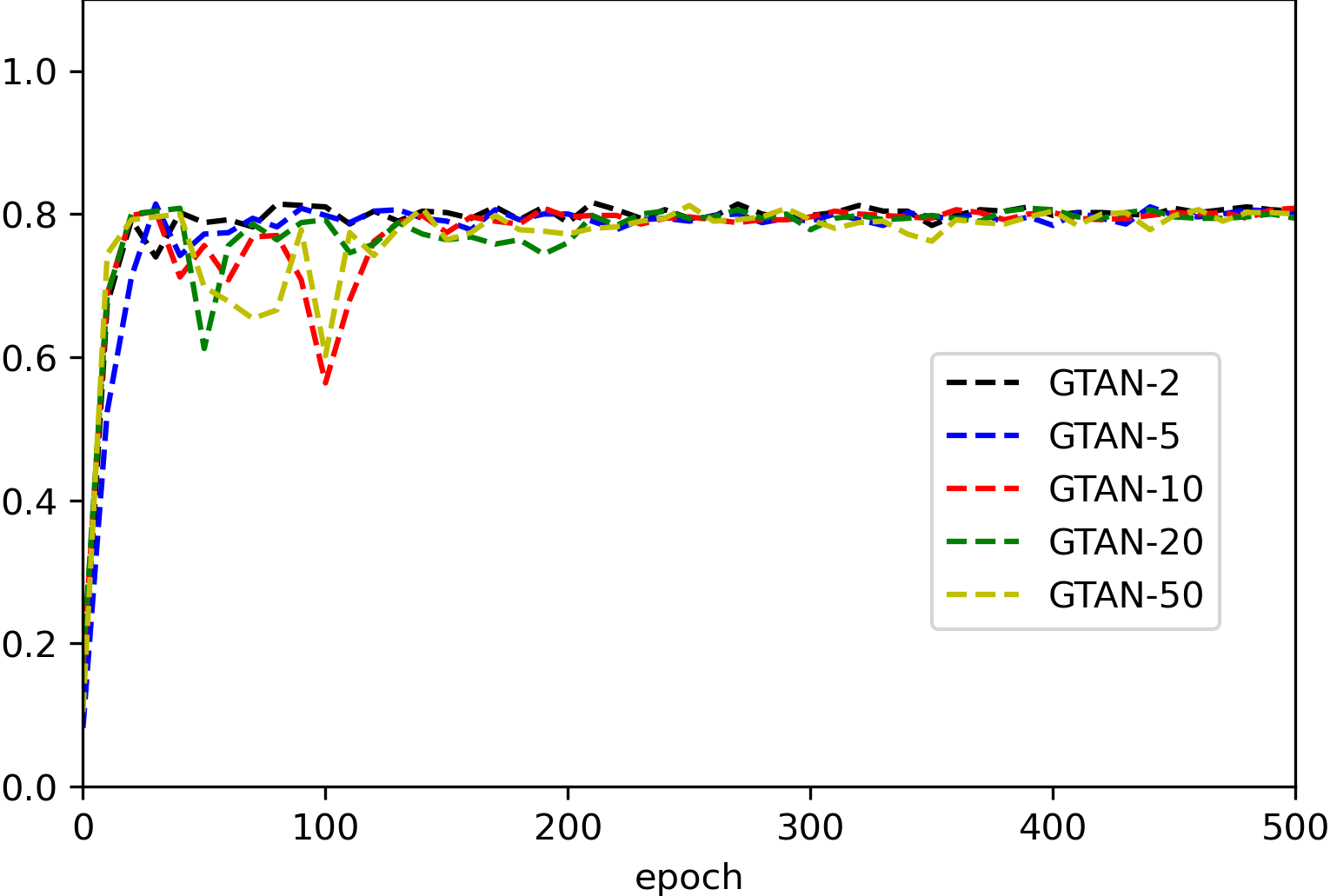}
			\caption{GTAN Validation Accuracy}
			\label{fig:GTAN_val}
		\end{center}
	\end{subfigure}
	\newline
	\begin{subfigure}[b]{0.235\textwidth}
		\begin{center}
			\includegraphics[width = 1.0\linewidth]{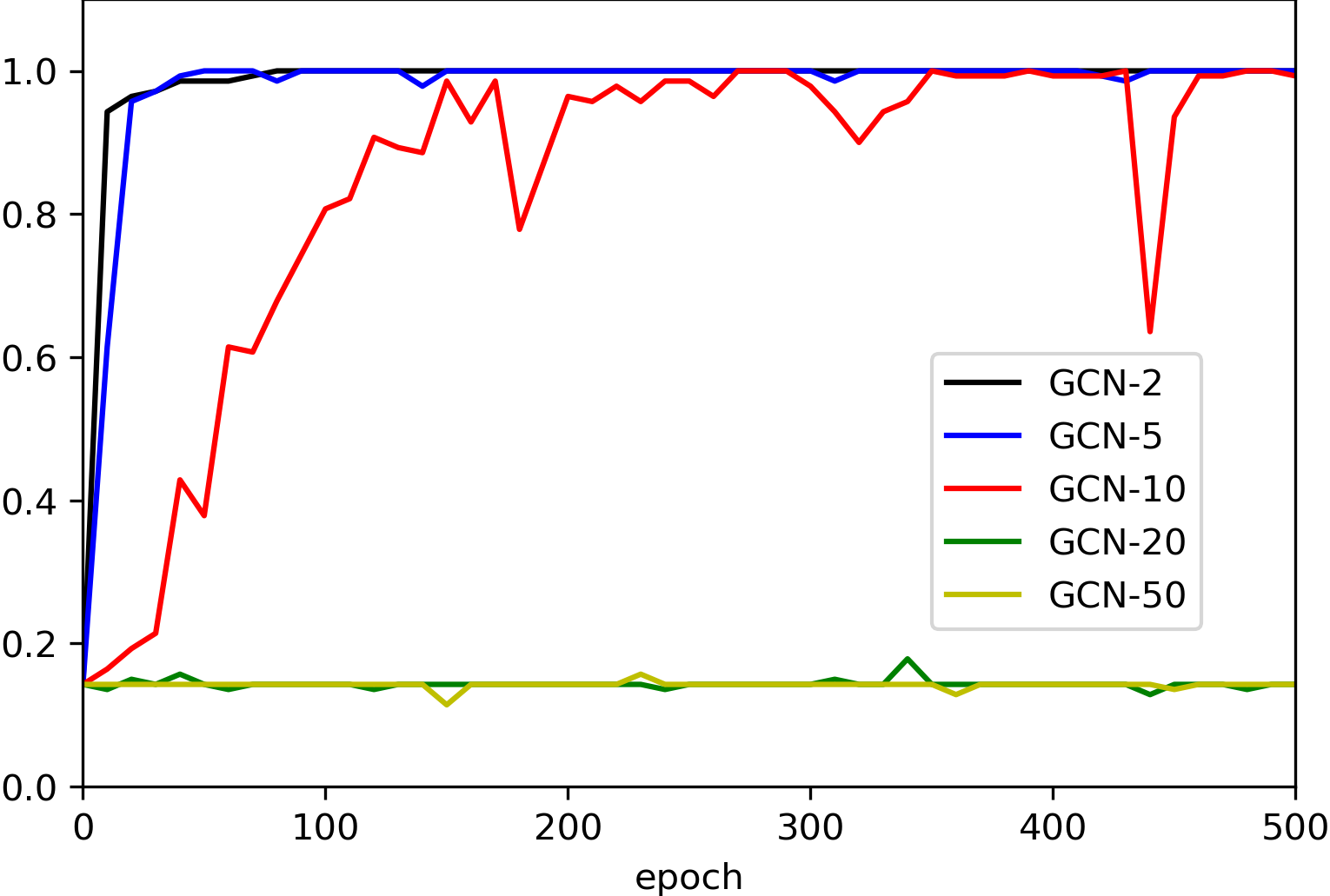}
			\caption{GCN Training Accuracy}
			\label{fig:GCN_train}
		\end{center}
	\end{subfigure}
	\begin{subfigure}[b]{0.235\textwidth}
		\begin{center}
			\includegraphics[width = 1.0\linewidth]{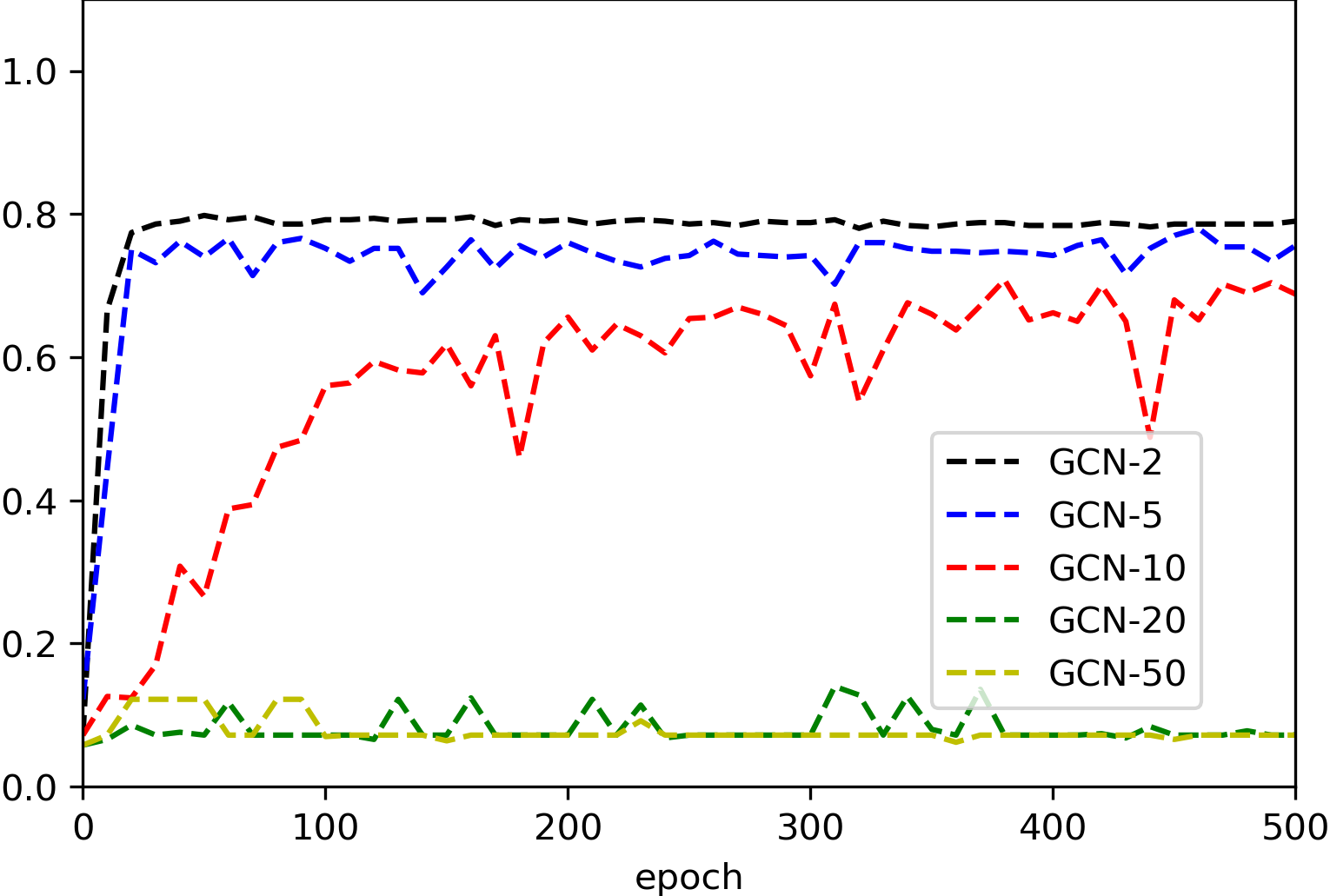}
			\caption{GCN Validation Accuracy}
			\label{fig:GCN_val}
		\end{center}
	\end{subfigure}
	\newline
	\begin{subfigure}[b]{0.235\textwidth}
		\begin{center}
			\includegraphics[width = 1.0\linewidth]{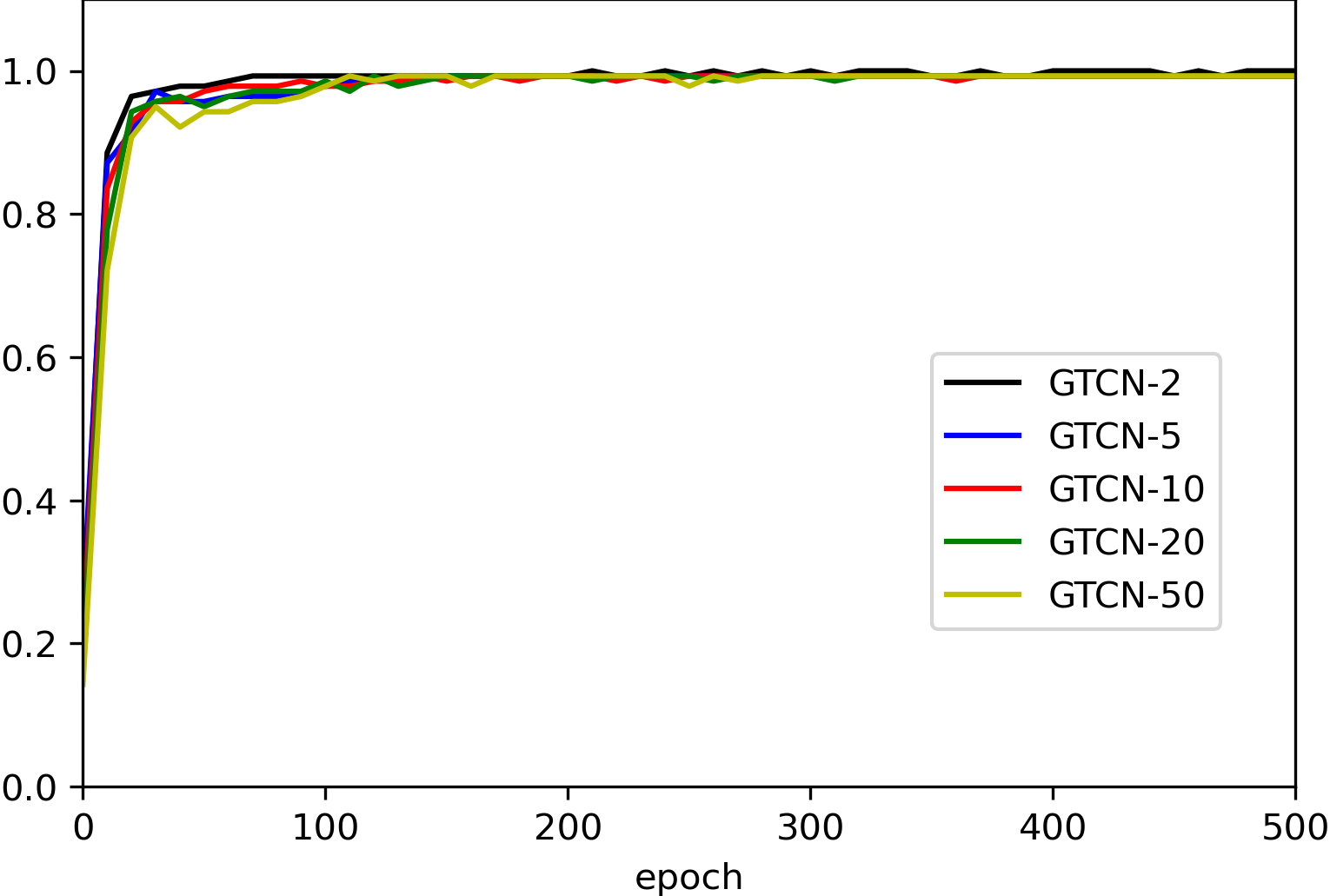}
			\caption{GTCN Training Accuracy}
			\label{fig:GTCN_train}
		\end{center}
	\end{subfigure}
	\begin{subfigure}[b]{0.235\textwidth}
		\begin{center}
			\includegraphics[width = 1.0\linewidth]{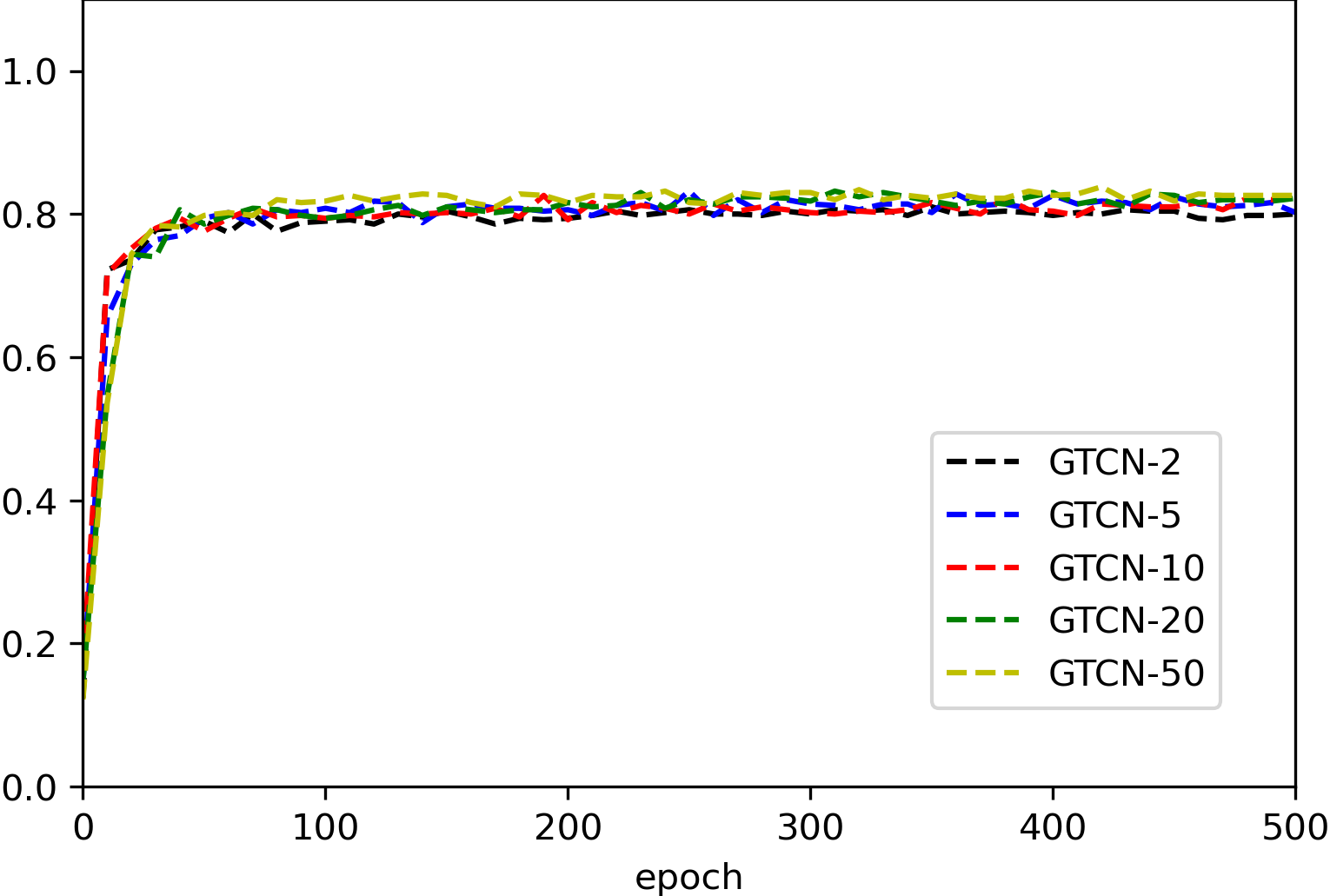}
			\caption{GTCN Validation Accuracy}
			\label{fig:GTCN_val}
		\end{center}
	\end{subfigure}
	\caption{Training and validation accuracy of GCN, GTCN, GAT and GTAN with different depths, tested on Cora dataset.}
	\label{fig:smooth}
\end{figure}

We look further into the over-smoothing issue by checking the training accuracy and validation accuracy over the number of epochs of the four models at different depths (2, 5, 10, 20, 50) on the Cora dataset. The results are displayed in \Cref{fig:smooth}. At depth of 20 and 50, both GAT and GCN stop responding to the samples due to the over-smoothing issue. The proposed GTAN and GTCN work consistently well at different depths.

\subsection{Ablation Study}
\label{ablation}
\begin{figure}[ht]
	\begin{subfigure}[b]{0.207\textwidth}
		\begin{center}
			\includegraphics[width = 1.0\linewidth]{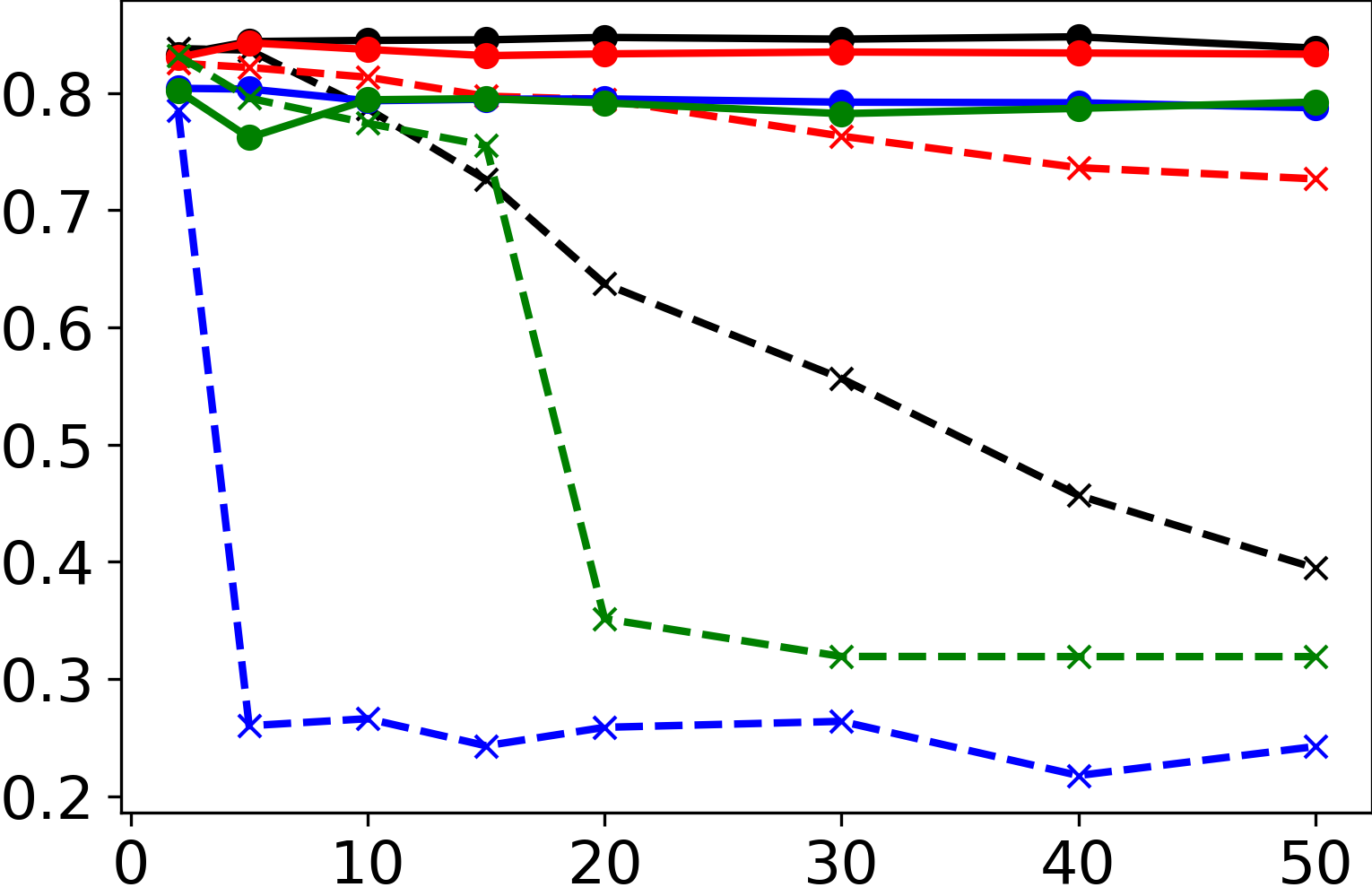}
			\caption{Cora}
			\label{fig:ablation:cora}
		\end{center}
	\end{subfigure}
	\hfill
	\begin{subfigure}[b]{0.269\textwidth}
		\begin{center}
			\includegraphics[width = 1.0\linewidth]{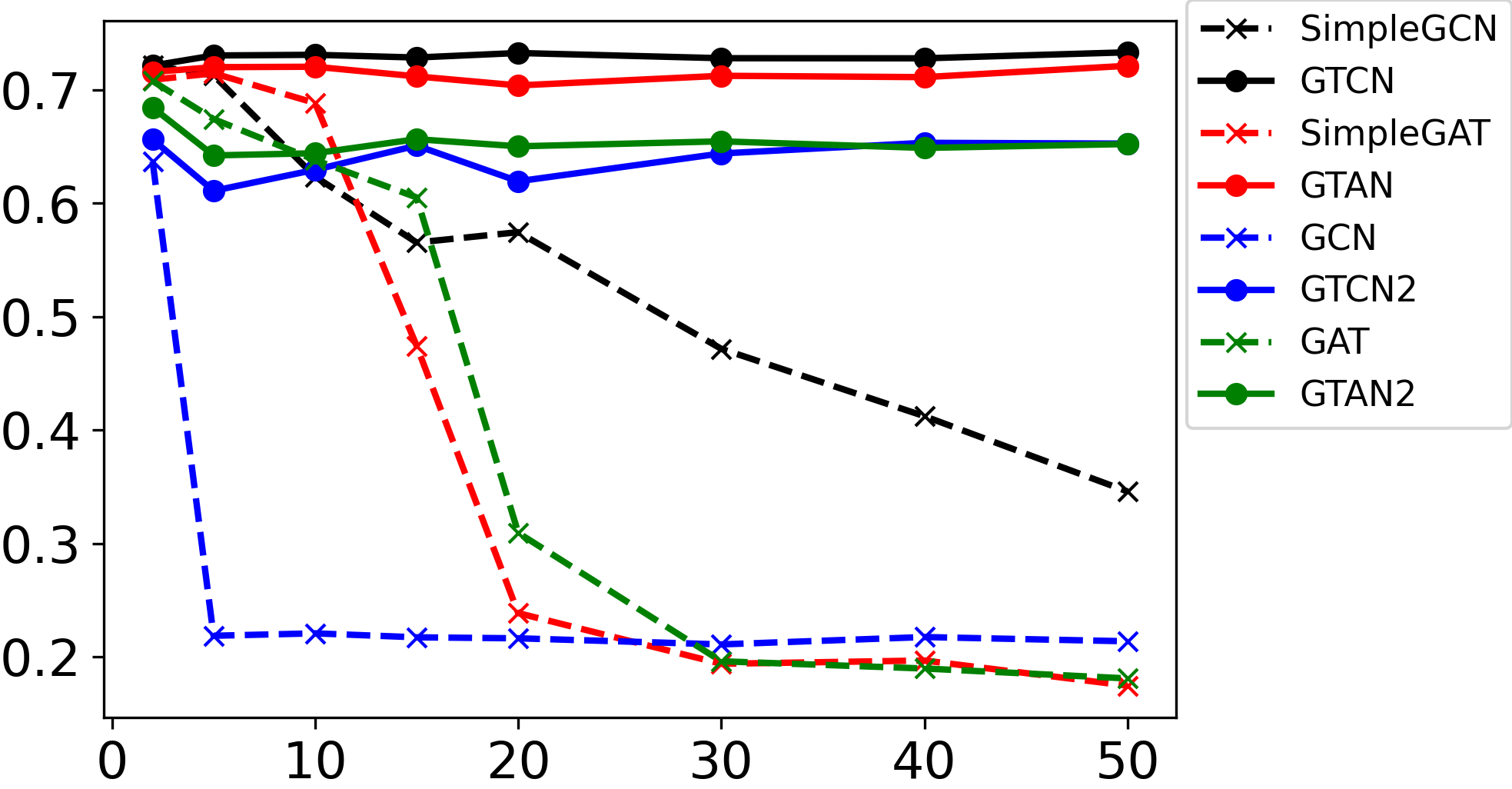}
			\caption{Citeseer}
			\label{fig:ablation:citeseer}
		\end{center}
	\end{subfigure}
	\newline
	\begin{subfigure}[b]{0.21\textwidth}
		\begin{center}
			\includegraphics[width = 1.0\linewidth]{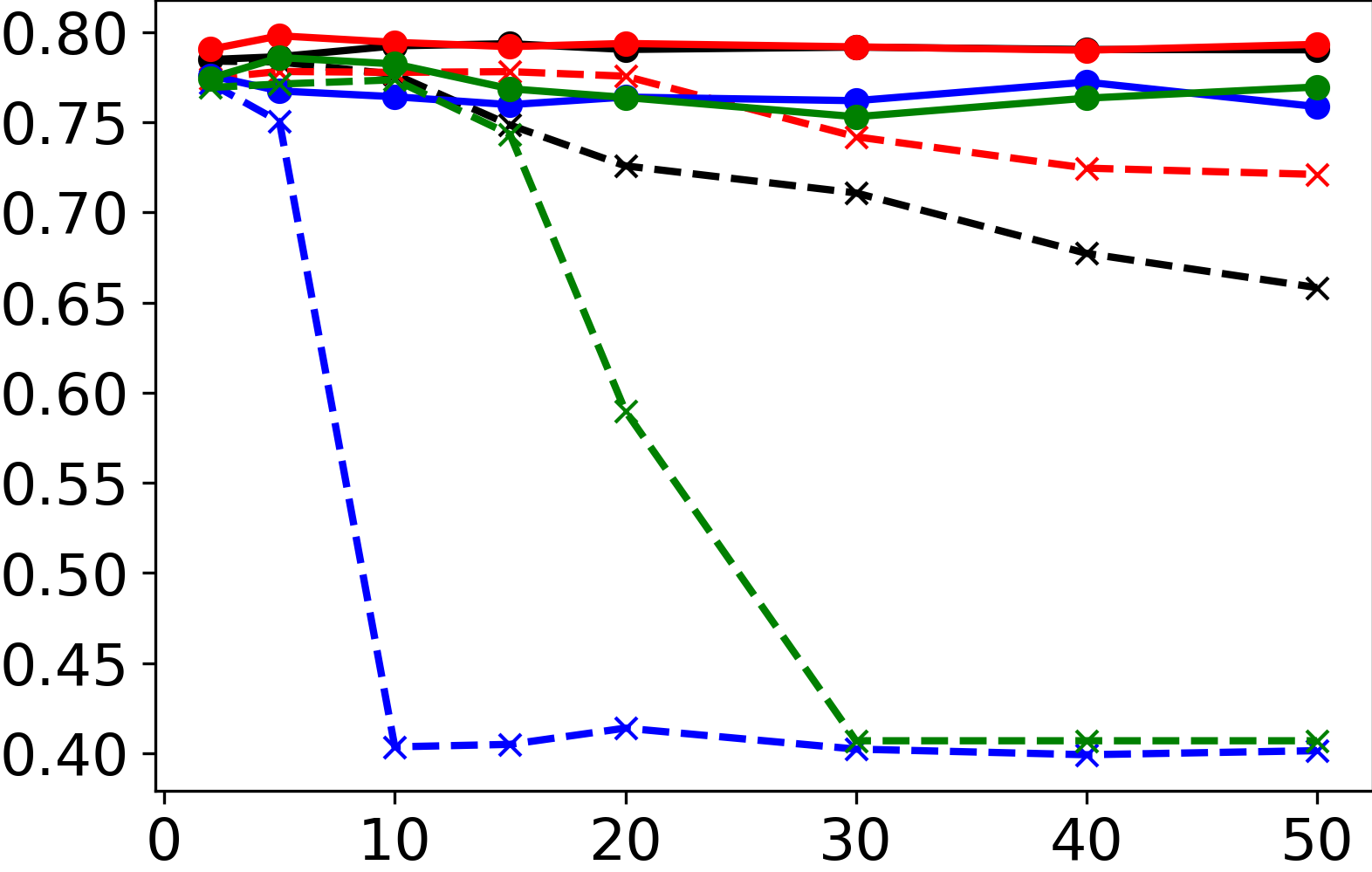}
			\caption{Pubmed}
			\label{fig:ablation:pubmed}
		\end{center}
	\end{subfigure}
	\hfill
	\begin{subfigure}[b]{0.265\textwidth}
		\begin{center}
			\includegraphics[width = 1.0\linewidth]{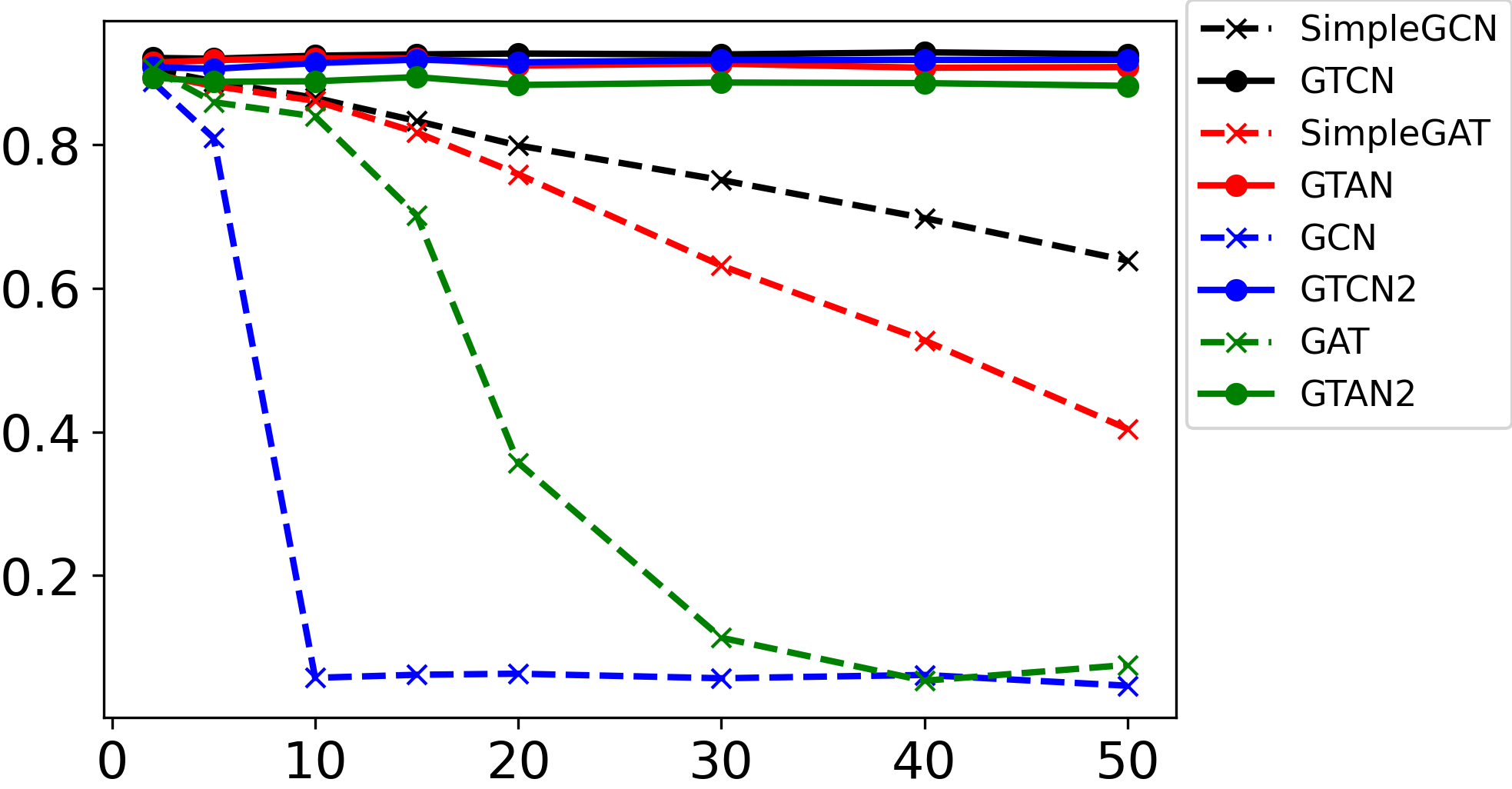}
			\caption{Coauthor-CS}
			\label{fig:ablation:coauthor}
		\end{center}
	\end{subfigure}
	\caption{Ablation Study}
	\label{fig:ablation}
\end{figure}

As discussed in \Cref{compare:GAT} and \ref{compare:GCN}, one of the major differences between GTAN and GAT, GTCN and GCN is that in each GAT and GCN layer, features are first transformed and then propagate. To examine the effect of the linear parameterized feature transformation, we build four variant models: simpleGAT and simpleGCN are based on the vanilla GAT and GCN by removing the linear feature transformation in each propagation layer, GTAN2 and GTCN2 are based on the proposed GTAN and GTCN by introducing the linear feature transformation in each propagation layer. We compare the average accuracy of the four models and their variants at different model depths on the four DGL datasets. The results are summarized in \Cref{fig:ablation}. It is seen that removing the linear feature transformation in GAT and GCN only slows down the over-smoothing problem, but does not resolve the problem. In contrast, adding a linear feature transformation in GTAN and GTCN may affect the model performance on small graphs due to over-fitting, but does not introduce the over-smoothing problem.

\section{Conclusion}
\label{conclusion}

In this paper, we propose a tree-based deep graph learning architecture: Graph Tree Network (GTNet), which is derived from the tree representation of graphs. The message passing scheme in GTNet follows the nature of message passing in the graph tree. Two models within the GTNet architecture are proposed - Graph Tree Attention Network (GTAN) and Graph Tree Convolution Network (GTCN), with demonstrated state-of-the-art performance on popular benchmark datasets and capability of going deep. The major advantage of the message passing scheme in GTNet is that models adopting this message passing scheme preserve nodes' local information, and can go deep by stacking multiple propagation layers instead of calculating multi-hop information directly. This advantage may allow us to extend our GTAN and GTCN models to heterogeneous graphs. Dealing with meta-paths in heterogeneous graphs is one of the biggest hurdles confronted by most of the current heterogeneous models like MAGNN \citep{fu2020magnn} and HAN \citep{wang2019heterogeneous}, as the number of meta-paths increases dramatically when aggregating multi-hop neighbors. The message passing scheme in GTNet allows a model to avoid dealing with metapaths explicity as the intra-metapath information is self-contained when stacking multiple propagation layers. This feature could lead to future research opportunities in heterogeneous graphs. The general message rule we formulate in the GTNet architecture may also open up new research opportunities by exploring combination of different aggregation schemes and transformation functions.

\section*{Acknowledgments}
This research is supported by the graduate fellowship from the Institute for Financial Services Analytics at University of Delaware.

\appendices
\begin{table*}[t]
	\caption{Average Macro-F1 score $\pm$ one standard deviation (in percent) on DGL datasets with the top and bottom 10\% data excluded.}
	\label{tb:F1}
	\begin{center}
		\begin{tabularx}{0.9\textwidth}{p{3.5cm} p{1.5cm} p{1.5cm} p{1.5cm} p{1.5cm} p{1.5cm} p{1.5cm}}
			\toprule
			\textbf{Method}  &\textbf{Cora} &\textbf{Citeseer}
			&\textbf{PubMed} &\textbf{Coauthor-CS (1)}
			&\textbf{Coauthor-CS (2)}
			&\textbf{Coauthor-CS (3)} \\
			\midrule
			GCN (hop = 2)  &80.5 $\pm$ 0.4 &68.8 $\pm$ 0.3 &78.8 $\pm$ 0.3 
			&87.7 $\pm$ 0.3 &87.5 $\pm$ 0.4 &86.4 $\pm$ 0.3\\
			GAT (hop = 2) &81.6 $\pm$ 0.5 &67.0 $\pm$ 0.7 &77.1 $\pm$ 0.3 &87.8 $\pm$ 0.4 &87.8 $\pm$ 0.4 &86.9 $\pm$ 0.4\\
			Tree-LSTM (hop = 2) &80.7 $\pm$ 0.6 &64.2 $\pm$ 0.9 &76.2 $\pm$ 0.6 &88.8 $\pm$ 0.2 &88.5 $\pm$ 0.2 &88.1 $\pm$ 0.2\\
			APPNP (hop = 10) &81.8 $\pm$ 0.4 &68.3 $\pm$ 0.3 &78.9 $\pm$ 0.2 &89.1 $\pm$ 0.5 &89.6 $\pm$ 0.3 &88.8 $\pm$ 0.5\\
			DAGNN (hop = 10) &82.5 $\pm$ 0.4 &68.9 $\pm$ 0.3 &78.9 $\pm$ 0.3 &87.5 $\pm$ 0.7 &87.4 $\pm$ 0.5 &86.9 $\pm$ 0.5\\
			\midrule
			GTAN (hop = 10, ours) &81.7 $\pm$ 0.8 &67.9 $\pm$ 0.8 &\textbf{79.0} $\pm$ \textbf{0.4} &90.0 $\pm$ 0.6 &90.3 $\pm$ 0.3 &89.9 $\pm$ 0.2\\
			GTCN (hop = 10, ours) &\textbf{82.8} $\pm$ \textbf{0.7} &\textbf{69.1} $\pm$ \textbf{0.6} &78.9 $\pm$ 0.3 &\textbf{90.6} $\pm$ \textbf{0.2} &\textbf{90.5} $\pm$ \textbf{0.2} &\textbf{90.2} $\pm$ \textbf{0.2}\\
			\bottomrule
		\end{tabularx}
	\end{center}
\end{table*}
\section{Proofs of Theorem and Lemma}
\label{proof}
\subsection{Proofs of Lemma \ref{lm:eigen1} and \ref{lm:eigen2}}
\begin{proof}
	For an arbitrary vector $\bm{x} \in \mathbb{R}^N$, we have
	
	\begin{align}
		\begin{split}
			\bm{x}^{T}\left(\bm{I}-\bm{A}_1\right)\bm{x} &= \sum_{i \in \mathcal{V}} x_i^2 - \sum_{(i,j)\in \mathcal{E}}\frac{x_i x_j}{\sqrt{(d_i+1)(d_j +1)}} \\
			&> \sum_{i \in \mathcal{V}} \frac{d_i}{d_i+1}x_i^2 - \sum_{(i,j)\in \mathcal{E}}\frac{x_i x_j}{\sqrt{(d_i+1)(d_j +1)}}\\
			&=\frac{1}{2}\sum_{(i,j)\in \mathcal{E}} \left(\frac{x_i}{\sqrt{d_i + 1}} - \frac{x_j}{\sqrt{d_j + 1}} \right)^2\\
			&\geq 0
		\end{split}
	\end{align}
	
	Therefore $\bm{I} - \bm{A}_1$ is positive definite, its eigenvalues are all positive, i.e. $\lambda_i > 0$ for $i=1,\ldots, N$.
	
	Further, we have
	\begin{equation}
		\frac{\bm{x}^{T} \bm{A}_1 \bm{x}}{\bm{x}^{T} \bm{x}} < 1 \Rightarrow a_i <1
	\end{equation}
	
	Similarly, we have
	
	\begin{align}
		\begin{split}
			\bm{x}^{T}\left(\bm{I}+\bm{A}_1\right)\bm{x} &= \sum_{i \in \mathcal{V}} x_i^2 + \sum_{(i,j)\in \mathcal{E}}\frac{x_i x_j}{\sqrt{(d_i+1)(d_j +1)}} \\
			&> \sum_{i \in \mathcal{V}} \frac{d_i}{d_i+1}x_i^2 + \sum_{(i,j)\in \mathcal{E}}\frac{x_i x_j}{\sqrt{(d_i+1)(d_j +1)}}\\
			&=\frac{1}{2}\sum_{(i,j)\in \mathcal{E}} \left(\frac{x_i}{\sqrt{d_i + 1}} + \frac{x_j}{\sqrt{d_j + 1}} \right)^2\\
			&\geq 0
		\end{split}
	\end{align}

Therefore we have 
\begin{equation}
	\frac{\bm{x}^{T} \bm{A}_1 \bm{x}}{\bm{x}^{T} \bm{x}} > -1 \Rightarrow a_i > -1
\end{equation}

We also have
\begin{align}
	\begin{split}
		\bm{x}^{T}\left(\bm{I}+\bm{A}_1\right)\bm{x} > 0 &\Rightarrow \bm{x}^{T}\left(\bm{I}-\bm{A}_1\right)\bm{x} < 2\bm{x}^{T}\bm{x} \\
		&\Rightarrow \frac{\bm{x}^{T}\left(\bm{I}-\bm{A}_1\right)\bm{x}}{\bm{x}^{T}\bm{x}} < 2\\
		& \Rightarrow \lambda_i < 2, \quad i = 1, \ldots, N
	\end{split}
\end{align}
\end{proof}

\subsection{Proof of Theorem \ref{thm:deep}}
\begin{proof}
	The message passing rule in GTCN is:
	\begin{align}
		\begin{split}
			&\bm{H}^k = \bm{A}_1 \bm{H}^{k+1} + \bm{A}_2 \bm{Z} \\
			&\bm{H}^L = \bm{Z}
		\end{split}
	\end{align}

The expression for the node representation at the final propagation layer can be derived iteratively as below:
\begin{align}
\begin{split}
	\bm{H}^0 &= \bm{A}_1 \bm{H}^1 + \bm{A}_2 \bm{Z} \\
	&= \bm{A}_1 \left(\bm{A}_1 \bm{H}^2 + \bm{A}_2 \bm{Z} \right) + \bm{A}_2 \bm{Z} \\
	&\quad \vdots \\
	&= \bm{A}_1^L \bm{H}^L + \left(\bm{A}_1^{L-1} + \bm{A}_1^{L-2} + \ldots + \bm{A}_1 + \bm{I}\right)\bm{Z}\\
	&= \left[\bm{A}_1^L + \left(\bm{I} - \bm{A}_1\right)^{-1} \left( \bm{I} - \bm{A}_1^L \right) \bm{A}_2 \right] \bm{Z}
\end{split}
\end{align}

From \textit{Lemma \ref{lm:eigen1}}, we have
\begin{equation}
	\lim\limits_{L \to \infty} a_i^L = 0
\end{equation}

Therefore
\begin{equation}
	\lim\limits_{L \to \infty} \bm{A}_1^L = \bm{0}
\end{equation}
and
\begin{equation}
	\lim\limits_{L \to \infty} \bm{H}^0 = \left(\bm{I} - \bm{A}_1\right)^{-1} \bm{A}_2 \bm{Z}
\end{equation}
\end{proof}

\subsection{Proof of Proposition 5}

\begin{proof}
	In a fully connected graph $\mathcal{G} = (\mathcal{V}, \mathcal{E})$, the normalized adjacency matrix is $\hat{\bm{A}} = (1/N)\mathbf{1}\mathbf{1}^T$, where $\hat{\bm{A}}_{uv} = 1/N, \forall u,v \in \mathcal{V}$.
	
	\textbf{GCN}. The hidden feature of node $u$ at the $n^\text{th}$ layer of GCN is updated as:
	\begin{equation}
		\label{eqn:GCN_eg}
		\bm{h}_u^n = \text{ReLU}\left(\frac{1}{N}\sum\nolimits_{v\in \mathcal{N}_u \cup \{u\}}{\bm{h}_v^{n-1} \bm{W}^n }\right)
	\end{equation}
	where $\bm{h}_u^0 = \bm{x}_u$.
	
	As $\sum\nolimits_{v\in \mathcal{N}_u \cup \{u\}}{\bm{h}_v^{n-1}}$ and the weight matrix $\bm{W}^n$ are the same for all nodes, the hidden features (i.e. node representation) of all nodes in a fully connected graph learned by GCN with any depth are identical.
	
	\textbf{GTCN}. The hidden feature of node $u$ at the $k^\text{th}$ hop of GTCN is updated as:
	\begin{equation}
		\label{eqn:gtcn_eg}
		\bm{h}_u^k = \frac{1}{N}\sum\nolimits_{v\in \mathcal{N}_u}{\bm{h}_v^{k+1}} + \frac{\bm{z}_u}{N}
	\end{equation}
	where $\bm{h}_u^L = \bm{z_u} = \text{MLP}(\bm{x}_u)$, $L$ is the total number of propagation layers (i.e. the size of neighborhood under consideration). Therefore we have
	\begin{equation*}
		\bm{h}_u^{L-1} = \frac{1}{N}\sum\nolimits_{v\in \mathcal{N}_u \cup \{u\}} \bm{z}_v
	\end{equation*}
	which is identical for all nodes. Let $\bm{h}^{L-1} \coloneqq \bm{h}_u^{L-1}, \forall u \in \mathcal{V}$, we have
	\begin{align*}
		\begin{split}
			&\bm{h}_u^{L-2}= \frac{N-1}{N} \bm{h}^{L-1} + \frac{\bm{z}_u}{N} \\
			&\bm{h}_u^{L-3}= \frac{(N-1)^2}{N^2} \bm{h}^{L-1} + \frac{1}{N^2}\sum\nolimits_{v\in \mathcal{N}_u}{\bm{z}_v} + \frac{\bm{z}_u}{N} \\
			&\vdots
		\end{split}
	\end{align*}

	The node representations learned by GTCN with more than one propagation layer on a fully connected graph preserves each node's local information $\bm{Z}$, and therefore all nodes remain distinguishable as shown in the above analysis.
\end{proof}

\section{Experimental Setup For Model Performance Demonstration}
\label{exp_setup}

In this section, we provide detailed experimental setups for reproducing our results. The number of maximum training epochs is set to 1000 for all experiments. Other hyperparameters are described in the following sections.

\subsection{DGL datasets}
For the vanilla GCN, we use two layers with 64 hidden units and the dropout is 0.5 for all four datasets. The learning rate and weight decay are 0.01 and 5e-4 respectively for all four datasets. 

For the vanilla GAT, we also use two layers with 64 hidden units, and other hyperparameters are fine-tuned to obtain the best performance. One attention head is used for both layers\footnote{The original paper \citep{DBLP:conf/iclr/VelickovicCCRLB18} uses 8 heads for the first layer, but we find that 1 head already achieves matching results.}. The layer dropout and attention dropout are 0.8 and 0.8 respectively for the Cora and Citeseer datasets, 0.8 and 0.2 respectively for the PubMed and Coauthor-CS datasets. The learning rate and weight decay are 0.01 and 5e-4 respectively for all four datasets. 

For the 2-layer Child-Sum Tree-LSTM model, the input dropout and the dropout after the LSTM cell are 0.8 and 0.6 respectively. The learning rate and weight decay are 0.01 and 5e-4 respectively for all four datasets. 

For the APPNP, DAGNN, and our GTCN and GTAN models, we use a maximum hop of 10, the same as that used by the original paper of the APPNP model \citep{klicpera2018predict}. 64 hidden units are used for these four deep models. For the APPNP, the dropout is 0.5 for the first MLP layer and the edge dropout is 0.5 for the propagation layer, the teleport probability $\alpha$ is 0.1 for the Cora, Citeseer and Pubmed datasets, and 0.2 for the Coauthor-CS dataset\footnote{We have validated with greedy search that the $\alpha$ value used in the original paper is approximately optimal.}. For the DAGNN, the dropout is 0.8 and learning rate is 0.01 for all four datasets. The weight decay is 2e-2 for the Citeseer dataset, and 5e-3 for the Cora, PubMed and Coauthor-CS datasets. For our GTCN, we have two dropouts: one is for the initial one-layer MLP, and the other is for the propagation layer. Corresponding dropout values are set to (0.6, 0.6) for the Cora dataset, (0.8, 0.6) for the Citeseer dataset, (0.8, 0.5) for the Pubmed dataset and (0.6, 0.2) for the Coauthor-CS dataset. The learning rate is 0.01 for the Cora, Citeseer and Coauthor-CS datasets, and 0.02 for the PubMed dataset. The weight decay is 5e-4 for the Cora, Citeseer and PubMed datasets, and 5e-3 for the Coauthor-CS dataset. For our GTAN, we have the same two dropouts as for the GTCN, which are set to (0.6, 0) for the Cora and Pubmed dataset, (0.6, 0.6) for the Citeseer dataset, and (0.2, 0.2) for the Coauthor-CS dataset. The learning rate is set to 0.01 for all four datasets. The weight decay is set to 5e-4 for the Cora, Citeseer and PubMed datasets, and 5e-3 for the Coauthor-CS dataset. 

The early-stopping patience number is set to 100 for the DAGNN model (the same number is used in the original paper \citep{liu2020towards}), 300 for our GTAN model, and 200 for all other models.

\subsection{OGB dataset}
For the vanilla GCN, we use three layers with 256 hidden units and the dropout is 0.5. The learning rate and weight decay are 0.01 and 0, respectively.

For the vanilla GAT, we use three layers with 128 hidden units. One attention head is used for all layers. The input dropout and attention dropout are 0.2 and 0, respectively. The learning rate and weight decay are 0.01 and 0, respectively.

For Child-Sum Tree-LSTM model, we use three layers with 256 hidden units. The input dropout and the dropout after the LSTM cell are 0.2 and 0, respectively. The learning rate and weight decay are 0.01 and 0, respectively. 

For the APPNP model, we use 5 layers with 256 hidden units. The input dropout and the dropout after each propagation layer are 0.2 and 0, respectively. The learning rate and weight decay are 0.01 and 0, respectively. 

For the DAGNN model, we use 16-hop with 256 hidden units. The dropout is 0.2. The learning rate and weight decay are 0.005 and 0, respectively. 

For our GTCN model, we use 5 layers with 256 hidden units. The input dropout and the dropout after each propagation layer are 0.2 and 0.2, respectively. The learning rate and weight decay are 0.01 and 5e-5, respectively.

For our GTAN model, we use 4 layers with 128 hidden units. The input dropout and the dropout after each propagation layer are 0.2 and 0, respectively. The learning rate and weight decay are 0.01 and 5e-5, respectively.

\section{Macro-F1 scores on DGL datasets}
\label{result:F1}
The average Macro-F1 scores on the four DGL datasets are summarized in \Cref{tb:F1}.



\bibliography{GTNet.bib}
\bibliographystyle{IEEEtran}

\end{document}